\documentclass[runningheads]{llncs}

\usepackage{eccv}

\usepackage{eccvabbrv}

\usepackage{graphicx}
\usepackage{booktabs}
\usepackage{graphicx}
\usepackage{multirow}
\usepackage{tabularx}
\usepackage{soul}
\usepackage{amsmath}
\DeclareUnicodeCharacter{2033}{''}
\usepackage[utf8]{inputenc}
\DeclareUnicodeCharacter{03BC}{$\mu$}

\usepackage[accsupp]{axessibility}  

\usepackage{hyperref}

\usepackage{orcidlink}
\usepackage{hyperref}

\usepackage{background}

\newcommand{\mywatermark}{%
    \begin{minipage}{\textwidth}
        \centering
        \fontsize{10}{10}\selectfont 
        This paper has been accepted for publication by the ECCV 2024 workshop on Neuromorphic Vision: Advantages and Applications of Event Cameras (NeVi).
    \end{minipage}%
}
\backgroundsetup{
    scale=1,
    color=gray,
    angle=0,
    opacity=0.5,
    position=current page.north,
    vshift=-1cm, 
    contents={\mywatermark}
}

\definecolor{highlight}{rgb}{0.9, 0.9, 0.9} 
\definecolor{highlight2}{rgb}{0.9, 0.9, 0.95} 
\definecolor{highlight3}{rgb}{0.95, 0.9, 0.9} 

\begin{document}


\title{Recent Event Camera Innovations: A Survey} 

\titlerunning{Event Cameras: A Survey}

\author{Bharatesh Chakravarthi\inst{1}\orcidlink{0000-0002-4978-434X
} \and
Aayush Atul Verma\inst{1}\orcidlink{0009-0007-4985-5404} \and
Kostas Daniilidis\inst{2}\orcidlink{0000-0003-0498-0758} \and
\\ Cornelia Fermuller\inst{3}\orcidlink{0000-0003-2044-2386}
\and
Yezhou Yang \inst{1}\orcidlink{0000-0003-0126-8976}
}

\authorrunning{B.~Chakravarthi et al.}

\institute{Arizona State University, Tempe AZ 85281, USA\\ 
\and
University of Pennsylvania, Philadelphia, PA 19104, USA\\
\and
University of Maryland, College Park, MD 20742, USA\\
}

\maketitle

\begin{abstract}
Event-based vision, inspired by the human visual system, offers transformative capabilities such as low latency, high dynamic range, and reduced power consumption. This paper presents a comprehensive survey of event cameras, tracing their evolution over time. It introduces the fundamental principles of event cameras, compares them with traditional frame cameras, and highlights their unique characteristics and operational differences. The survey covers various event camera models from leading manufacturers, key technological milestones, and influential research contributions. It explores diverse application areas across different domains and discusses essential real-world and synthetic datasets for research advancement. Additionally, the role of event camera simulators in testing and development is discussed. This survey aims to consolidate the current state of event cameras and inspire further innovation in this rapidly evolving field. To support the research community, a \href{https://github.com/chakravarthi589/Event-based-Vision_Resources}{GitHub} page categorizes past and future research articles and consolidates valuable resources.

  
\end{abstract}

\begin{figure}[t]
\centerline{\includegraphics[width=0.70\linewidth]{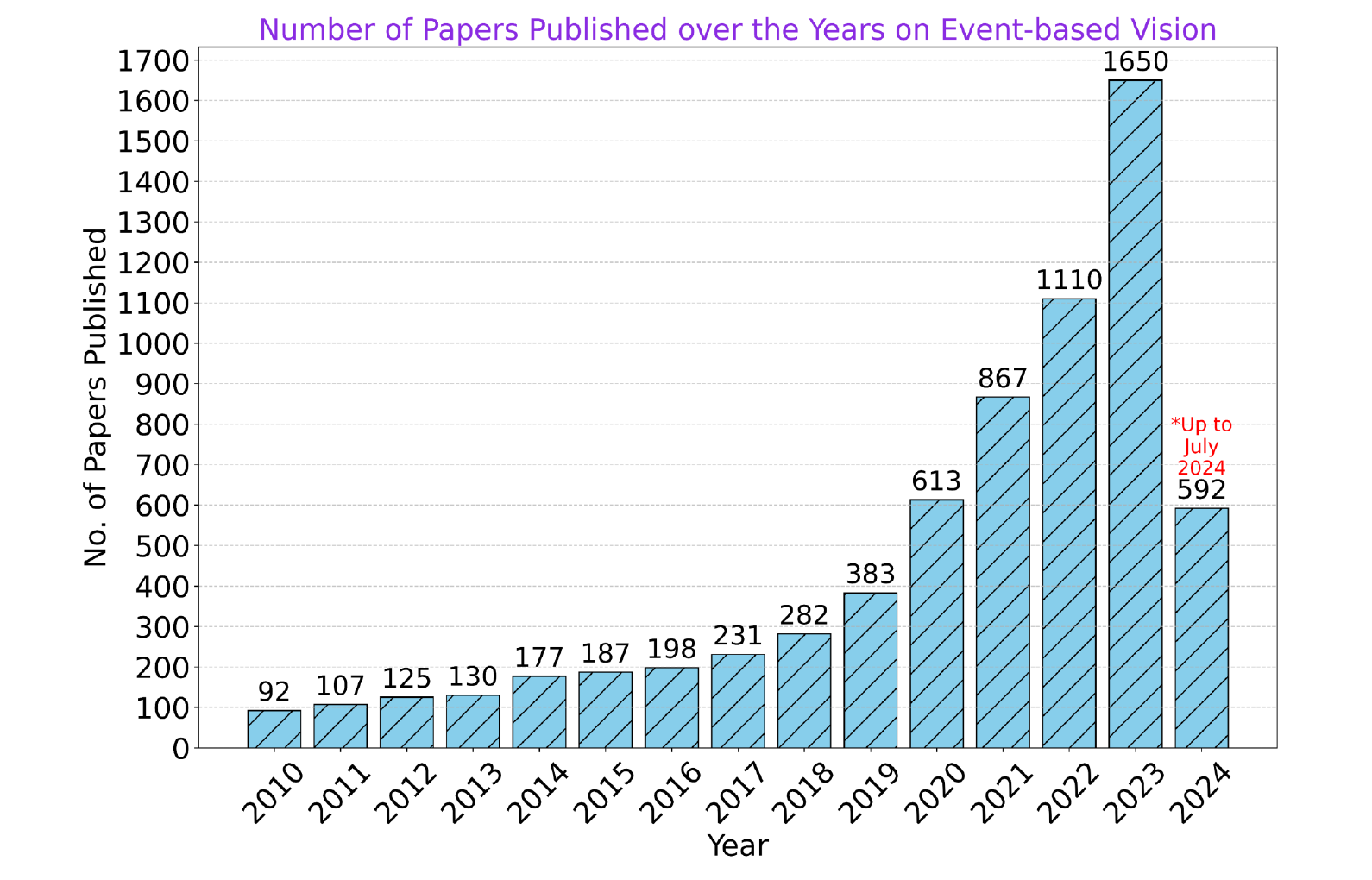}}
\caption{Publication Trends in Event-based Vision Research.}
\label{fig01}
\end{figure}

\section{Understanding Event-based Vision -- An Introduction }
\label{Intro}

Event-based vision represents a paradigm shift in visual sensing technology, inspired by the human visual system’s capability (thus also referred to as neuromorphic vision) to detect and respond to changes in the environment. Unlike traditional frame cameras, which capture static images at set intervals, event-based vision utilizes event cameras to continuously monitor light intensity changes at each pixel. These cameras produce \textbf{``events''} only when significant changes occur, generating a dynamic data stream that reflects real-time scene dynamics. 
Event-based vision mimics the asynchronous nature of human perception, where each pixel independently detects and records changes. This approach provides exceptionally high temporal resolution, essential for accurately capturing fast-moving objects and dynamic scenes without the motion blur typically associated with frame cameras. By focusing solely on changes and excluding static information, event cameras manage data more efficiently, significantly reducing redundancy and bandwidth requirements.

The real-time capture and processing of events enable immediate responses to scene changes, making event-based vision particularly suitable for applications that require rapid decision-making. The technology’s focus on detecting changes on a logarithmic scale rather than absolute values allows it to handle a wide range of lighting conditions effectively, avoiding common issues like overexposure or underexposure encountered with traditional systems. This adaptability is particularly valuable in environments with challenging lighting conditions in outdoor settings. 
Moreover, since event cameras process only the changes, they require less data bandwidth and computational power compared to traditional cameras. This efficiency results in significant energy savings, making event-based vision ideal for battery-powered devices and long-term monitoring applications. The asynchronous nature also facilitates efficient data handling and analysis, focusing exclusively on relevant changes and enabling faster more accurate processing. 

The distinctive features of event cameras - low latency, high dynamic range, reduced power consumption, and efficient data handling have driven their adoption across various application domains, including object detection \cite{gehrig2023recurrent}, moving object segmentation \cite{mitrokhin2018event,mitrokhin2019ev,mitrokhin2020learning,parameshwara20210}, object tracking \cite{zhang2022spiking,  ramesh2018long}, object classification \cite{sironi2018hats, bi2019graph}, gesture\slash action recognition \cite{liu2022fast, amir2017low,deniz2023neuromorphic}, flow\slash depth\slash pose estimation \cite{benosman2013event, zhu2018ev, zhu2019unsupervised, mueggler2017event, mueggler2014event}, semantic segmentation \cite{alonso2019ev, sun2022ess}, video deblurring \cite{jiang2020learning, lin2020learning}, video generation \cite{wang2019event, liu2017high}, neural radiance fields (NERF) \cite{klenk2023nerf, rudnev2023eventnerf}, visual odometry \cite{censi2014low, zihao2017event, zhou2021event,ye2020unsupervised}, high-resolution video reconstruction \cite{chen2024timerewind,zhang2022unifying,tulyakov2022time} and motion capture \cite{xu2020eventcap, iaboni2022event, millerdurai2024eventego3d}.

This survey aims to provide researchers with a comprehensive understanding of the current state of event cameras. It offers a background on research trends to illustrate the increasing interest in this field (\cref{StatStudy}). The survey explains the workings of event cameras (\cref{eventCamWorking}) and contrasts them with traditional frame cameras (\cref{eventVsFrame}). 
It studies various event camera models from leading manufacturers, providing a feature-wise comparison to aid in camera selection (\cref{camTypes}). Key milestone works are overviewed to set the stage for future research directions (\cref{eventMilestones}). Additionally, the survey discusses the diverse application areas of event-based vision, presenting notable works across different domains (\cref{eventApplications}). An overview of crucial event-based datasets (\cref{eventDatasets}) and simulators essential for advancing research and development is also included (\cref{eventSimulators}). 

The objective of this survey is to consolidate event-based vision systems resources, emphasizing technological advancements and practical applications while serving as a thorough guide to the field's features and options. Complementing this survey is a 
\href{https://github.com/chakravarthi589/Event-based-Vision_Resources}{GitHub } resource page, which will be regularly updated to provide researchers with the latest developments in event-based vision, facilitating informed decision-making and promoting ongoing innovation.

\section{The Rise of Event-based Vision: A Background}
\label{StatStudy}

The event-based vision research community has seen significant advancements in recent years, as evidenced by the growing number of published papers (see \cref{fig01}). Starting from a modest number in $2010$, the field has expanded, peaking with a substantial increase in scholarly activity by $2023$. This notable surge, particularly from $2019$ onwards, is attributed to the increased availability of event cameras from various vendors and the introduction of advanced event-based simulators. Major computer vision conferences such as CVPR, ECCV, ICCV, and WACV have observed a significant rise in event-based vision research papers. For instance, the number of event-based vision papers presented at CVPR has grown markedly, from a few in $2018$ to a considerable number in $2024$. Specialized workshops dedicated to event-based vision have further contributed to disseminating research in this area. This trend illustrates its expanding impact and increasing recognition within the broader computer vision community.

In the late $1990s$ and early $2000s$, notable advancements in neuromorphic vision included the development of a neuromorphic sensor for robots \cite{ harrison1998neuromorphic}, spiking neural controllers \cite{ floreano2001evolution}, biologically inspired vision sensors \cite{van2002biologically} and an open-source toolkit for neuromorphic vision \cite{itti2004ilab}. Key works also featured a review of artificial human vision technologies \cite{dowling2005artificial}, an embedded real-time tracking system \cite{litzenberger2006embedded} and a multichip system for spike-based processing \cite{vogelstein2007multichip}. Additionally, the frame-free dynamic digital vision was discussed \cite{delbruck2008frame}, AER dynamic vision sensors were introduced for a balancing robot \cite{conradt2009embedded, conradt2009pencil}, a dynamic stereo-vision system was developed \cite{schraml2010dynamic}, and activity-driven sensors were introduced \cite{delbruck2010activity}. Notably, \cite{bulthoff2003biologically} organized a workshop on biologically motivated vision.

From the early $2010s$ to $2020$, notable advancements included the exploration of asynchronous event-based binocular stereo-matching \cite{rogister2011asynchronous}, embedded neuromorphic vision for humanoid robots \cite{bartolozzi2011embedded}, multi-kernel convolution processor module \cite{camunas2011event}, high-speed vision for microparticle tracking \cite{ni2012asynchronous}, temporally correlated features extraction \cite{bichler2012extraction, lichtsteiner2008128}, and an algorithm for recognition \cite{maashri2012accelerating}. Researchers advanced techniques in event-based visual flow \cite{benosman2013event}, SLAM \cite{weikersdorfer2013simultaneous}\slash $3D$ SLAM \cite{weikersdorfer2014event}, a robotic goalie with rapid reaction times \cite{delbruck2013robotic}, and a review on retinomorphic sensors \cite{posch2014retinomorphic}.  Additionally, a multi-kernel algorithm for high-speed visual features tracking \cite{lagorce2014asynchronous}, continuous-time trajectory estimation \cite{mueggler2015continuous}, lifetime estimation of events and visual tracking  \cite{mueggler2015lifetime}, stereo matching  \cite{firouzi2016asynchronous}, and spiking neural network model of $3D$ perception \cite{osswald2017spiking} were explored during the mid-$2010s$. Innovations such as event-driven classifier \cite{stromatias2017event}, spatiotemporal filter for reducing noise \cite{khodamoradi2018n}, low-latency line tracking \cite{everding2018low}, graph-based object classification \cite{bi2019graph}, and gait recognition \cite{wang2019ev} emerged. The late $2010s$ saw comprehensive surveys on event-based vision \cite{gallego2020event} and neuromorphic vision for autonomous driving \cite{chen2020event} along with spatiotemporal feature learning for neuromorphic vision sensing \cite{bi2020graph}. The rapid advent of event cameras and simulators in the early $2020s$ significantly impacted the field, leading to milestone achievements as discussed in \cref{eventMilestones}.  

\begin{figure}[t]
\centerline{\includegraphics[width=0.90
\linewidth]{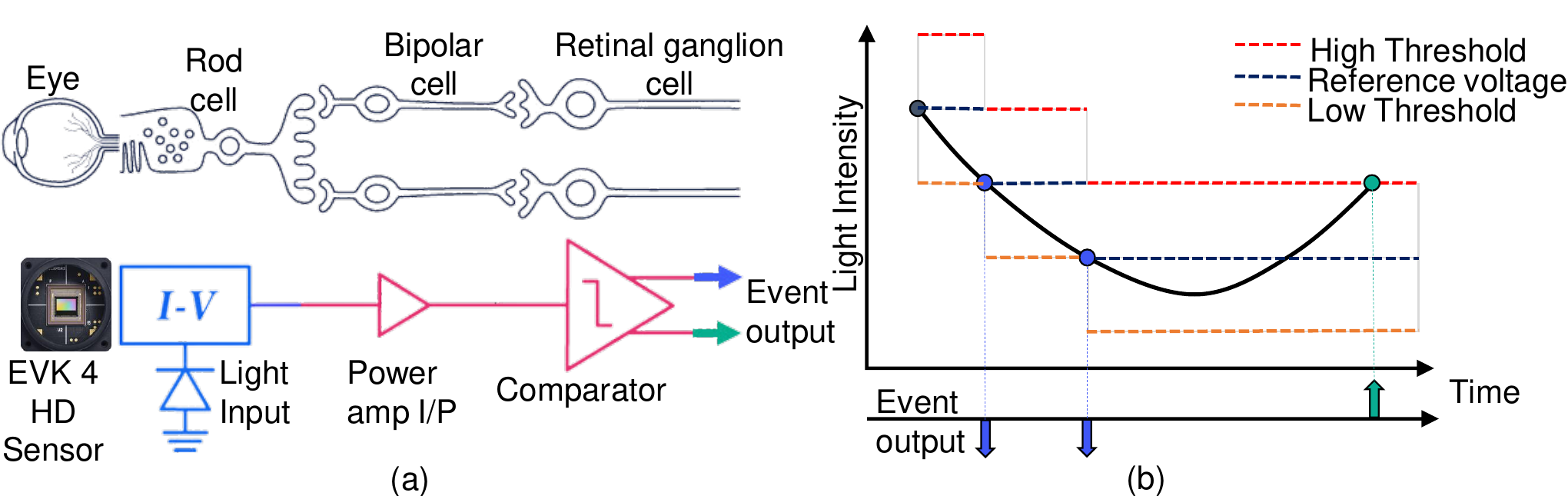}}
\caption{Working Mechanism of Event Cameras: (a) Independent pixel operation converting light into voltage signals for detecting intensity changes. (b) Event generation as a function of logarithmic light intensity over time.}
\label{fig02}
\end{figure}

\section{How Event Cameras Work: An Inside Look}
\label{eventCamWorking}
Event-based vision fundamentally differs from traditional frame-based vision in the way they process a scene. Inspired by the human retina, where the rod, bipolar, and retinal ganglion cells detect and transmit visual signals independently (see \cref{fig02}~(a)), the purpose of each pixel in the sensor is to capture any change in visual information of the scene asynchronously. This autonomous principle of the sensor leads to a unique and efficient way of processing visual data in real-time.
The working mechanism of an event camera involves several key steps. Each pixel operates independently, continuously, and asynchronously processing incoming light. Light photons hitting photodiodes in each pixel are converted into electrical current and transformed into a voltage signal. This generated voltage is compared to a reference voltage at each pixel continuously to detect logarithmic changes in light intensity. 

As illustrated in Fig.~\ref{fig02}~(b), every time the voltage difference exceeds a predefined threshold, an event $\langle x, y, p, t \rangle$ is triggered, recording the pixel coordinates $(x, y)$, the time of change $t$, and the polarity $p\in\{-1, +1\}$, denoting an increase or decrease of light intensity. These events are output as they occur, reflecting the changes in the scene over time through a continuous data stream rather than a series of static frames. The stream can be visualized as a two-channel representation in a $3D$ space. Here, two dimensions constitute the spatial component capturing the location of the event in image coordinates and the third dimension represents its temporal coordinates, indicating precisely when the event occurred. This spatial-temporal representation minimizes data redundancy and enables efficient processing of dynamic aspects of the scene through its sparse structure.

\begin{figure}[t]
\centerline{\includegraphics[width=0.8\linewidth]{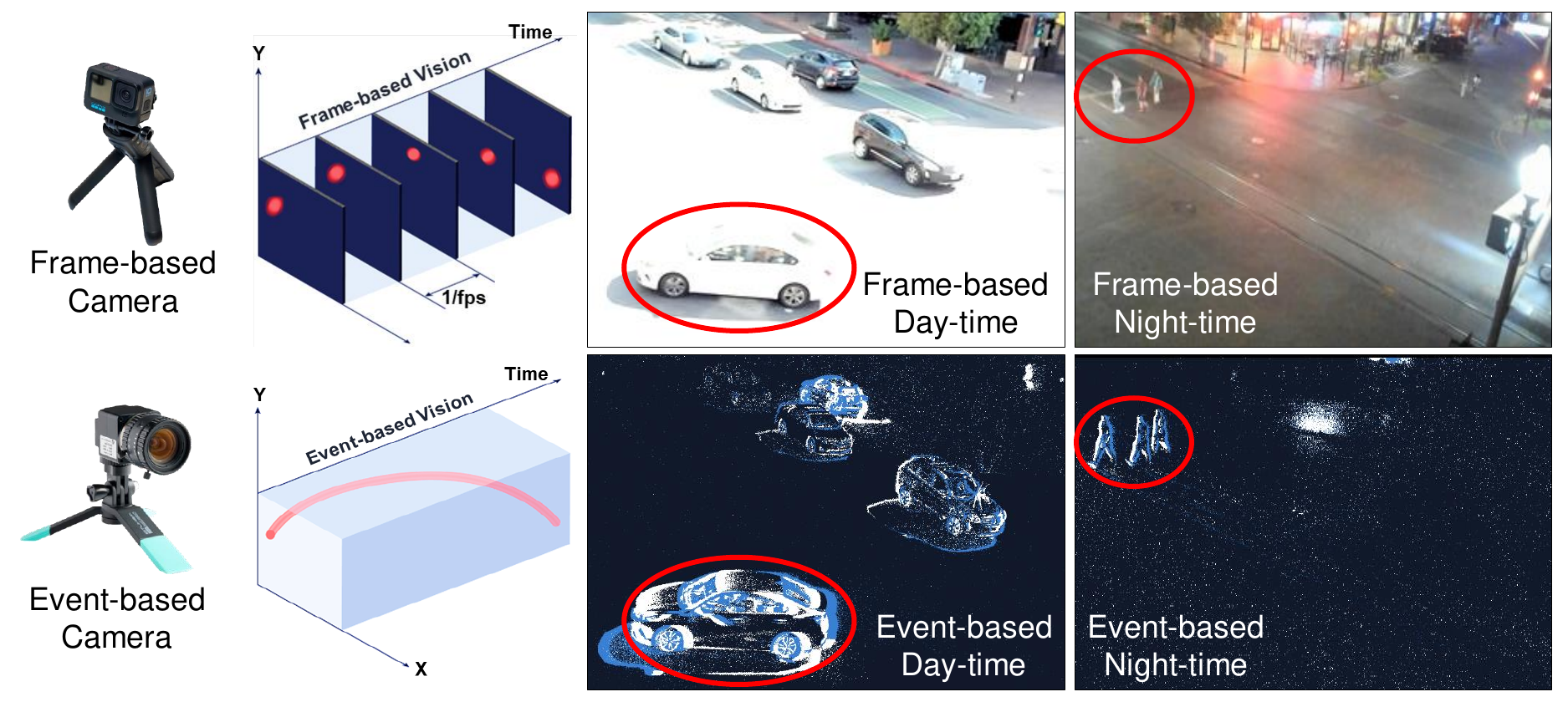}}
\caption{Comparison of Frame vs. Event Cameras: The top row shows common issues like motion blur and visibility in frame-based images, while the bottom row shows event-based images with reduced motion blur and better visibility in challenging lighting.}
\label{fig03}
\end{figure}

\section{Event Camera vs. Frame Camera: A Comparison}
\label{eventVsFrame}
Event cameras offer several advantages over traditional frame cameras due to their unique operating principles. Each pixel in an event camera records changes the moment they are detected, allowing for the capture of fast-moving objects and dynamic scenes, thus achieving high temporal resolution (\textgreater$10,000\,\text{fps}$). Motion blur, a common issue in frame-based systems \cite{Dai_2023_CVPR}, occurs when objects move rapidly during the camera's exposure time, causing them to appear smeared across the image. However, unlike frame cameras, where every pixel must wait for a global exposure time of the frame, event cameras respond immediately to changes in the scene. This immediate response helps event cameras demonstrate low latency and significantly reduce motion blur, as shown in~\cref{fig03}. This capability is critical in applications requiring real-time monitoring and rapid response, such as robotics and autonomous driving~\cite{9252186, falanga2020dynamic}. 

While modern frame cameras can achieve high frame rates, they come at the cost of large bandwidth and storage requirements, which can be limiting. By only recording changes in the scene, event cameras produce less data than traditional frame cameras. This reduction in data bandwidth makes event cameras ideal for applications with limited bandwidth or storage capacity. The focus on changes rather than absolute light levels further ensures that only the relevant information is captured, reducing redundancy. These advantages are most important for embedded and edge device systems, which often have limited processing power, memory, and storage capabilities, and benefit greatly from efficient and reduced data output~\cite{10067520, sridharan2024ev, safa2022neuromorphic, gokarn2024poster, liang2024towards}.

Furthermore, event cameras operate effectively across a wide range of lighting conditions. Focusing on logarithmic changes in light intensity allows them to avoid issues like overexposure, underexposure, and sudden changes in lighting conditions that commonly affect traditional cameras. Event sensors'  high dynamic range (\textgreater$120\;\text{dB}$) far exceeds the dynamic range exhibited by high-quality frame cameras that do not exceed $95\;\text{dB}$~\cite{imatestDynamicRange}. This makes them suitable for environments with challenging lighting (see Fig.~\ref{fig03}), such as outdoor scenes with varying illumination. Their exceptional low light cutoff ($0.08\,\text{Lux}$) has prompted further explorations in various low-light applications~\cite{zhang2020learning, 9278914, mahlknecht2022exploring}. Overall, these advantages make event cameras a compelling choice for diverse applications.

\section{Event Camera Models: An Overview}
\label{camTypes}
In 2017, pioneering works \cite{ amir2017low,  mueggler2017event} utilized early event cameras like the DVS 128 \cite{ inivation_dvs128} and DAVIS 240 \cite{ inivation_davis240}, laying the groundwork for advanced applications in the field. Since then, event camera technology has advanced significantly, with prominent manufacturers such as iniVation \cite{inivation_website}, Prophesee \cite{ prophesee_website}, Lucid Vision Lab (TRT009S-EC, TRT003S-EC) \cite{lucid_website}, Celepixe (CeleX5-MIP, CeleX-V), and Insightness (SiliconEye Rino 3 EVK) \cite{insightness_website} introducing innovative event camera models. 
Among these, iniVation and Prophesee have emerged as leaders, with models such as the DAVIS 346 \cite{davis346_event_camera}, Prophesee EVK4 \cite{metavision_evK4_HD}, and DAVIS 240 \cite{ inivation_davis240}  gaining prominence in the research community. This section reviews various event cameras from iniVation and Prophesee.

iniVation is a leading company in neuromorphic vision systems, known for its bio-inspired technology that provides ultra-low latency, high dynamic range, and low power consumption. Their current product lineup includes the DVXplorer \cite{ dvxplorer_event_camera} with VGA resolution, 110 dB dynamic range, and 165 million events per second, the DVXplorer Lite \cite{dvxplorer_lite_event_camera} with QVGA resolution, 110 dB dynamic range, and 100 million events per second, the DAVIS 346 \cite{davis346_event_camera}, a prototype offering concurrent QVGA\texttt{+} resolution and up to 12 million events per second, and the DAVIS 346 AER \cite{ davis_346_aer_event_camera}, which provides both event and frame output with a 120 dB dynamic range. Moreoever, the DVXplorer S Duo \cite{ dvxplorer_s_duo_event_camera} integrates an event-based sensor with a global-shutter color image sensor, powered by an Nvidia Jetson Nano SOM. Also, their Stereo Kit \cite{ stereo_kit} includes two devices, lenses, tripods, and other accessories for advanced stereo vision exploration. Note that some earlier products, such as the DVXplorer Mini, DVS 240, DAVIS 240, eDVS, DVS 128, DVL-5000, have been discontinued by iniVation and are no longer available. Additionally, iniVation offers software solutions like DV \cite{ dv_software} for user-friendly visualization, DV-Processing \cite{dv_processing} for \texttt{C++}/\texttt{Python}-based processing, and ROS integration, alongside a low-level library for event camera usage. \cref{inivationCams} summarizes the key characteristics and features of iniVation’s event cameras.

\begin{table*}[t]
\centering
\caption{Comprehensive Comparison of Event Cameras Manufactured by iniVation.}
\label{inivationCams}
\resizebox{0.875\textwidth}{!}{%
\begin{tabular}{|llcccccc}
\multicolumn{2}{l}{\multirow{2}{*}{}} &
  \includegraphics[width=2.4cm]{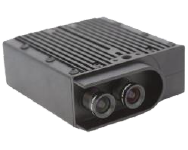}  &
  \includegraphics[width=2cm]{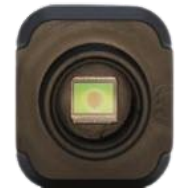} &
  \includegraphics[width=2.2cm]{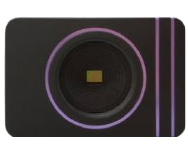} &
  \includegraphics[width=2.2cm]{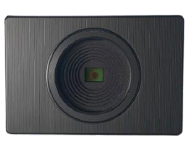} &
  \includegraphics[width=2.2cm]{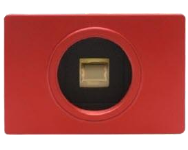} &
  \includegraphics[width=2.4cm]{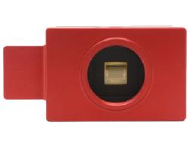} \\ \cline{3-8}  
\multicolumn{2}{l}{} &
  \multicolumn{1}{|c|}{\textbf{DVXplorer S Duo}} &
  \multicolumn{1}{c|}{\textbf{DVXplorer Micro}} &
  \multicolumn{1}{c|}{\textbf{DVXplorer}} &
  \multicolumn{1}{c|}{\textbf{DVXplorer Lite}} &
  \multicolumn{1}{c|}{\textbf{DAVIS346}} &
  \multicolumn{1}{c|}{\textbf{DAVIS346 AER}} \\ \hline \hline
  
\multirow{7}{*}{\rotatebox[origin=c]{90}{\centering \textbf{Event Output}}} 

&
  \multicolumn{1}{|l|}{\textbf{Spatial Resolution}} &
  \multicolumn{1}{c|}{640 x 480} &
  \multicolumn{1}{c|}{640 x 480} &
  \multicolumn{1}{c|}{640 x 480} &
  \multicolumn{1}{c|}{320 x 240} &
  \multicolumn{1}{c|}{320 x 240} &
  \multicolumn{1}{c|}{320 x 240} \\ \cline{2-8} 
\multicolumn{1}{|l|}{} &
  \multicolumn{1}{l|}{\textbf{Temporal Resolution}} &
  \multicolumn{1}{c|}{65 - 200 $\mu$s} &
  \multicolumn{1}{c|}{65 - 200 $\mu$s} &
  \multicolumn{1}{c|}{65 - 200 $\mu$s} &
  \multicolumn{1}{c|}{65 - 200 $\mu$s} &
  \multicolumn{1}{c|}{1 $\mu$s} &
  \multicolumn{1}{c|}{1 $\mu$s} \\ \cline{2-8} 
\multicolumn{1}{|l|}{} &
  \multicolumn{1}{l|}{\textbf{Max Throughput}} &
  \multicolumn{1}{c|}{30 MEPS} &
  \multicolumn{1}{c|}{450 MEPS} &
  \multicolumn{1}{c|}{165 MEPS} &
  \multicolumn{1}{c|}{100 MEPS} &
  \multicolumn{1}{c|}{12 MEPS} &
  \multicolumn{1}{c|}{12 MEPS} \\ \cline{2-8} 
\multicolumn{1}{|l|}{} &
  \multicolumn{1}{l|}{\textbf{Latency}} &
  \multicolumn{1}{c|}{\textless{}1 ms} &
  \multicolumn{1}{c|}{\textless{}1 ms} &
  \multicolumn{1}{c|}{\textless{}1 ms} &
  \multicolumn{1}{c|}{\textless{}1 ms} &
  \multicolumn{1}{c|}{\textless{}1 ms} &
  \multicolumn{1}{c|}{\textless{}1 ms} \\ \cline{2-8} 
\multicolumn{1}{|l|}{} &
  \multicolumn{1}{l|}{\textbf{Dynamic Range}} &
  \multicolumn{1}{c|}{90 dB - 110 dB} &
  \multicolumn{1}{c|}{90 dB - 110 dB} &
  \multicolumn{1}{c|}{90 dB - 110 dB} &
  \multicolumn{1}{c|}{90 dB - 110 dB} &
  \multicolumn{1}{c|}{120 dB} &
  \multicolumn{1}{c|}{120 dB} \\ \cline{2-8} 
\multicolumn{1}{|l|}{} &
  \multicolumn{1}{l|}{\textbf{Contrast Sensitivity}} &
  \multicolumn{1}{c|}{13\% - 27.5\%} &
  \multicolumn{1}{c|}{13\% - 27.5\%} &
  \multicolumn{1}{c|}{13\% - 27.5\%} &
  \multicolumn{1}{c|}{13\% - 27.5\%} &
  \multicolumn{1}{c|}{14.3\% - 22.5\%} &
  \multicolumn{1}{c|}{14.3\% - 22.5\%} \\ \cline{2-8} 
\multicolumn{1}{|l|}{} &
  \multicolumn{1}{l|}{\textbf{Pixel pitch}} &
  \multicolumn{1}{c|}{9 $\mu$m} &
  \multicolumn{1}{c|}{9 $\mu$m} &
  \multicolumn{1}{c|}{9 $\mu$m} &
  \multicolumn{1}{c|}{18 $\mu$m} &
  \multicolumn{1}{c|}{18.5 $\mu$m} &
  \multicolumn{1}{c|}{18.5 $\mu$m} \\ \hline
  \multirow{7}{*}{\rotatebox[origin=c]{90}{\centering \textbf{Frame Output}}}

&
  \multicolumn{1}{|l|}{\textbf{Spatial Resolution}} &
  \multicolumn{1}{c|}{1920 x 1080 HD} &
  \multicolumn{1}{c|}{-} &
  \multicolumn{1}{c|}{-} &
  \multicolumn{1}{c|}{-} &
  \multicolumn{1}{c|}{346 x 260} &
  \multicolumn{1}{c|}{346 x 260} \\ \cline{2-8} 
\multicolumn{1}{|l|}{} &
  \multicolumn{1}{l|}{\textbf{Frame Rate}} &
  \multicolumn{1}{c|}{Up to 30 fps} &
  \multicolumn{1}{c|}{-} &
  \multicolumn{1}{c|}{-} &
  \multicolumn{1}{c|}{-} &
  \multicolumn{1}{c|}{Up to 40 fps} &
  \multicolumn{1}{c|}{Up to 40 fps} \\ \cline{2-8} 
\multicolumn{1}{|l|}{} &
  \multicolumn{1}{l|}{\textbf{Dynamic Range}} &
  \multicolumn{1}{c|}{71.4 dB} &
  \multicolumn{1}{c|}{-} &
  \multicolumn{1}{c|}{-} &
  \multicolumn{1}{c|}{-} &
  \multicolumn{1}{c|}{55 dB} &
  \multicolumn{1}{c|}{55 dB} \\ \cline{2-8} 
\multicolumn{1}{|l|}{} &
  \multicolumn{1}{l|}{\textbf{FPN}} &
  \multicolumn{1}{c|}{-} &
  \multicolumn{1}{c|}{-} &
  \multicolumn{1}{c|}{-} &
  \multicolumn{1}{c|}{-} &
  \multicolumn{1}{c|}{4.2 \%} &
  \multicolumn{1}{c|}{4.2 \%} \\ \cline{2-8} 
\multicolumn{1}{|l|}{} &
  \multicolumn{1}{l|}{\textbf{Dark Signal}} &
  \multicolumn{1}{c|}{-} &
  \multicolumn{1}{c|}{-} &
  \multicolumn{1}{c|}{-} &
  \multicolumn{1}{c|}{-} &
  \multicolumn{1}{c|}{18000 e– /s} &
  \multicolumn{1}{c|}{18000 e– /s} \\ \cline{2-8} 
\multicolumn{1}{|l|}{} &
  \multicolumn{1}{l|}{\textbf{Readout Noise}} &
  \multicolumn{1}{c|}{-} &
  \multicolumn{1}{c|}{-} &
  \multicolumn{1}{c|}{-} &
  \multicolumn{1}{c|}{-} &
  \multicolumn{1}{c|}{55 e-} &
  \multicolumn{1}{c|}{55 e-} \\ \cline{2-8} 
\multicolumn{1}{|l|}{} &
  \multicolumn{1}{l|}{\textbf{Pixel Pitch}} &
  \multicolumn{1}{c|}{3 $\mu$m} &
  \multicolumn{1}{c|}{-} &
  \multicolumn{1}{c|}{-} &
  \multicolumn{1}{c|}{-} &
  \multicolumn{1}{c|}{18.5 $\mu$m} &
  \multicolumn{1}{c|}{18.5 $\mu$m} \\ \hline

\multirow{9}{*}{\rotatebox[origin=c]{90}{\centering \textbf{Device Attributes}}} 

&
  \multicolumn{1}{|l|}{\textbf{Dimensions (H x W x D)}} &
  \multicolumn{1}{c|}{32 x 80 x 92} &
  \multicolumn{1}{c|}{24 x 27.5 x 29.7} &
  \multicolumn{1}{c|}{40 x 60 x 25} &
  \multicolumn{1}{c|}{40 x 60 x 25} &
  \multicolumn{1}{c|}{40 x 60 x 25} &
  \multicolumn{1}{c|}{40 x 78.8 x 25} \\ \cline{2-8} 
\multicolumn{1}{|l|}{} &
  \multicolumn{1}{l|}{\textbf{Lens mount}} &
  \multicolumn{1}{c|}{S-mount} &
  \multicolumn{1}{c|}{CS-mount} &
  \multicolumn{1}{c|}{CS-mount} &
  \multicolumn{1}{c|}{CS-mount} &
  \multicolumn{1}{c|}{CS-mount} &
  \multicolumn{1}{c|}{CS-mount} \\ \cline{2-8} 
\multicolumn{1}{|l|}{} &
  \multicolumn{1}{l|}{\textbf{Mounting options}} &
  \multicolumn{1}{c|}{\begin{tabular}[c]{@{}c@{}}2- side Whitworth \\ 1/4″- 20 female and \\ M3 mounting points\end{tabular}} &
  \multicolumn{1}{c|}{4x M2 mounting points} &
  \multicolumn{4}{c|}{4-side Whitworth 1/4″-20 female and M3 mounting points} \\ \cline{2-8} 
\multicolumn{1}{|l|}{} &
  \multicolumn{1}{l|}{\textbf{Connectors}} &
  \multicolumn{1}{c|}{USB 3.0 C, Mini HDMI} &
  \multicolumn{1}{c|}{USB 3.0 micro B port} &
  \multicolumn{1}{c|}{USB 3.0 micro B port} &
  \multicolumn{1}{c|}{USB 3.0 micro B port} &
  \multicolumn{1}{c|}{USB 3.0 micro B port} &
  \multicolumn{1}{c|}{USB 3.0 micro B port} \\ \cline{2-8} 
\multicolumn{1}{|l|}{} &
  \multicolumn{1}{l|}{\textbf{Case materials}} &
  \multicolumn{1}{c|}{Anodized aluminum} &
  \multicolumn{1}{c|}{Engineering plastic} &
  \multicolumn{1}{c|}{Anodized aluminum} &
  \multicolumn{1}{c|}{Engineering plastic} &
  \multicolumn{1}{c|}{Anodized aluminum} &
  \multicolumn{1}{c|}{Anodized aluminum} \\ \cline{2-8} 
\multicolumn{1}{|l|}{} &
  \multicolumn{1}{l|}{\textbf{Weight (without lense)}} &
  \multicolumn{1}{c|}{220 g} &
  \multicolumn{1}{c|}{16 g} &
  \multicolumn{1}{c|}{100 g} &
  \multicolumn{1}{c|}{75 g} &
  \multicolumn{1}{c|}{100 g} &
  \multicolumn{1}{c|}{120 g} \\ \cline{2-8} 
\multicolumn{1}{|l|}{} &
  \multicolumn{1}{l|}{\textbf{Power consumption}} &
  \multicolumn{1}{c|}{7W – 12 W} &
  \multicolumn{4}{c|}{\textless 140 mA @ 5 VDC (USB)} &
  \multicolumn{1}{c|}{\textless{}180 mA @ 5 VDC (USB)} \\ \cline{2-8} 
\multicolumn{1}{|l|}{} &
  \multicolumn{1}{l|}{\textbf{Sensor Technology}} &
  \multicolumn{5}{c|}{90 nm BSI CIS} &
  \multicolumn{1}{c|}{0.18 $\mu$m 1P6M MIM CIS} \\ \cline{2-8} 
\multicolumn{1}{|l|}{} &
  \multicolumn{1}{l|}{\textbf{Sensor Supply voltage}} &
  \multicolumn{5}{c|}{1.2 V, 1.8 V and 2.8 V} &
  \multicolumn{1}{c|}{1.8 V and 3.3 V} \\ \hline
\multicolumn{2}{|c|}{\multirow{3}{*}{\textbf{Other Features}}} &
  \multicolumn{6}{c|}{6-axis IMU  (Gyro \texttt{+} Accelerometer), up to 8 kHz sampling rate} \\ \cline{3-8} 
\multicolumn{2}{|c|}{} &
  \multicolumn{1}{c|}{-} &
  \multicolumn{1}{c|}{-} &
  \multicolumn{3}{c|}{Supports multi-camera time synchronization} &
  \multicolumn{1}{c|}{-} \\ \cline{3-8} 
\multicolumn{2}{|c|}{} &
  \multicolumn{1}{c|}{\begin{tabular}[c]{@{}c@{}}on-board processing\\ (Nvidia Jetson Nano)\end{tabular}} &
  \multicolumn{1}{c|}{-} &
  \multicolumn{1}{c|}{-} &
  \multicolumn{1}{c|}{-} &
  \multicolumn{1}{c|}{-} &
  \multicolumn{1}{c|}{-} \\ \hline \hline
\end{tabular}%
}
\end{table*}

\begin{table*}[h]
\centering
\caption{Comprehensive Comparison of Event Cameras Manufactured by Prophesee.}
\label{propheseeCams}
\resizebox{0.875\textwidth}{!}{%
\begin{tabular}{|llcccccc}
\multicolumn{2}{l}{\multirow{2}{*}{}} &
  \includegraphics[width=2.4cm]{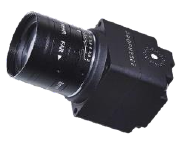} &
  \includegraphics[width=2.4cm]{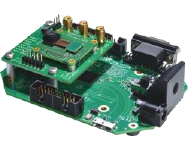} &
  \includegraphics[width=2.4cm]{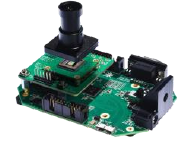} &
  \includegraphics[width=2.4cm]{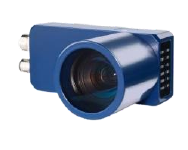} &
  \includegraphics[width=2.4cm]{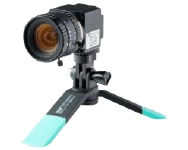} &
  \includegraphics[width=2.4cm]{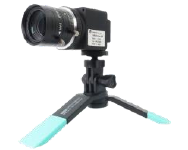} \\ \cline{3-8} 
\multicolumn{2}{l}{} &
  \multicolumn{1}{|c|}{\textbf{EVK 4 HD}} &
  \multicolumn{1}{c|}{\textbf{EVK3}} &
  \multicolumn{1}{c|}{\textbf{EVK3 - HD}} &
  \multicolumn{1}{c|}{\textbf{VisionCam EB}} &
  \multicolumn{1}{c|}{\textbf{SilkyEvCam VGA}} &
  \multicolumn{1}{c|}{\textbf{SilkyEvCam HD}} \\ \hline \hline

\multirow{8}{*}{\rotatebox[origin=c]{90}{\centering \textbf{Event Output}}} 
  
 &
  \multicolumn{1}{|l|}{\textbf{Spatial Resolution}} &
  \multicolumn{1}{c|}{1280 x 720} &
  \multicolumn{1}{c|}{320 x 320} &
  \multicolumn{1}{c|}{1280 x 720} &
  \multicolumn{1}{c|}{640 x 480} &
  \multicolumn{1}{c|}{640 x 480} &
  \multicolumn{1}{c|}{1280 x 720} \\ \cline{2-8} 
\multicolumn{1}{|l|}{} &
  \multicolumn{1}{l|}{\textbf{Optical Format}} &
  \multicolumn{1}{c|}{1/2.5"} &
  \multicolumn{1}{c|}{1/5"} &
  \multicolumn{1}{c|}{1/5"} &
  \multicolumn{1}{c|}{3/4"} &
  \multicolumn{1}{c|}{3/4"} &
  \multicolumn{1}{c|}{1/2.5"} \\ \cline{2-8} 
\multicolumn{1}{|l|}{} &
  \multicolumn{1}{l|}{\textbf{Max. Bandwidth}} &
  \multicolumn{1}{c|}{1.6 Gbps} &
  \multicolumn{1}{c|}{-} &
  \multicolumn{1}{c|}{1.6 Gbps} &
  \multicolumn{1}{c|}{1 Gbps} &
  \multicolumn{1}{c|}{-} &
  \multicolumn{1}{c|}{-} \\ \cline{2-8} 
\multicolumn{1}{|l|}{} &
  \multicolumn{1}{l|}{\textbf{Typical Latency}} &
  \multicolumn{1}{c|}{220 $\mu$s} &
  \multicolumn{1}{c|}{\textless{}150 $\mu$s} &
  \multicolumn{1}{c|}{220 $\mu$s} &
  \multicolumn{1}{c|}{220 $\mu$s} &
  \multicolumn{1}{c|}{200 $\mu$s} &
  \multicolumn{1}{c|}{\textless 100 $\mu$s} \\ \cline{2-8} 
\multicolumn{1}{|l|}{} &
  \multicolumn{1}{l|}{\textbf{Dynamic Range}} &
  \multicolumn{1}{c|}{\textgreater 86 dB} &
  \multicolumn{1}{c|}{\textgreater 120 dB} &
  \multicolumn{1}{c|}{\textgreater{}86 dB} &
  \multicolumn{1}{c|}{\textgreater{}120 dB} &
  \multicolumn{1}{c|}{\textgreater 120 dB} &
  \multicolumn{1}{c|}{\textgreater 120 dB} \\ \cline{2-8} 
\multicolumn{1}{|l|}{} &
  \multicolumn{1}{l|}{\textbf{Contrast Sensitivity}} &
  \multicolumn{1}{c|}{25\%} &
  \multicolumn{1}{c|}{25\%} &
  \multicolumn{1}{c|}{25\%} &
  \multicolumn{1}{c|}{25\%} &
  \multicolumn{1}{c|}{25\%} &
  \multicolumn{1}{c|}{25\%} \\ \cline{2-8} 
\multicolumn{1}{|l|}{} &
  \multicolumn{1}{l|}{\textbf{Pixel Size}} &
  \multicolumn{1}{c|}{4.86 x 4.86 $\mu$m} &
  \multicolumn{1}{c|}{6.3 x 6.3 $\mu$m} &
  \multicolumn{1}{c|}{4.86 x 4.86 $\mu$m} &
  \multicolumn{1}{c|}{15 x 15 $\mu$m} &
  \multicolumn{1}{c|}{15 x 15 $\mu$m} &
  \multicolumn{1}{c|}{4.86 x 4.86 $\mu$m} \\ \cline{2-8} 
\multicolumn{1}{|l|}{} &
  \multicolumn{1}{l|}{\textbf{Low light cutoff}} &
  \multicolumn{1}{c|}{0.08 lux} &
  \multicolumn{1}{c|}{0.05 lux} &
  \multicolumn{1}{c|}{0.08 lux} &
  \multicolumn{1}{c|}{0.08 lux} &
  \multicolumn{1}{c|}{0.08 lux} &
  \multicolumn{1}{c|}{0.08 lux} \\ \hline

\multirow{9}{*}{\rotatebox[origin=c]{90}{\centering \textbf{Device Attributes}}} 

&
  \multicolumn{1}{|l|}{\textbf{\begin{tabular}[c]{@{}l@{}}Dimensions\\ (H x W x D) mm\end{tabular}}} &
  \multicolumn{1}{c|}{30 x 30 x 36} &
  \multicolumn{1}{c|}{108 x 76 x 45} &
  \multicolumn{1}{c|}{108 x 76 x 45} &
  \multicolumn{1}{c|}{\begin{tabular}[c]{@{}c@{}}105 × 50  × \\ (30 \texttt{+} lens tube 27 – 80 mm)\end{tabular}} &
  \multicolumn{1}{c|}{30 x 30 x 36} &
  \multicolumn{1}{c|}{30 x 30 x 36} \\ \cline{2-8} 
\multicolumn{1}{|l|}{} &
  \multicolumn{1}{l|}{\textbf{D-FoV}} &
  \multicolumn{1}{c|}{47.7$^\circ$} &
  \multicolumn{1}{c|}{76$^\circ$} &
  \multicolumn{1}{c|}{81.5$^\circ$} &
  \multicolumn{1}{c|}{-} &
  \multicolumn{1}{c|}{70$^\circ$} &
  \multicolumn{1}{c|}{47.7$^\circ$} \\ \cline{2-8} 
\multicolumn{1}{|l|}{} &
  \multicolumn{1}{l|}{\textbf{Lens Mount}} &
  \multicolumn{1}{c|}{C/CS mount} &
  \multicolumn{1}{c|}{M12 S-Mount} &
  \multicolumn{1}{c|}{C/CS with S-mount} &
  \multicolumn{1}{c|}{C-Mount} &
  \multicolumn{1}{c|}{C/CS Mount} &
  \multicolumn{1}{c|}{C/CS Mount} \\ \cline{2-8} 
\multicolumn{1}{|l|}{} &
  \multicolumn{1}{l|}{\textbf{Mounting Options}} &
  \multicolumn{1}{c|}{\begin{tabular}[c]{@{}c@{}}4x M2 front \texttt{+} 2x M2.6 \\ back fixing points\end{tabular}} &
  \multicolumn{1}{c|}{\begin{tabular}[c]{@{}c@{}}Optical Flex module \\ / COB module\end{tabular}} &
  \multicolumn{1}{c|}{M12 S-Mount} &
  \multicolumn{1}{c|}{6 × M4} &
  \multicolumn{1}{c|}{M12 S-Mount} &
  \multicolumn{1}{c|}{M12 S-Mount} \\ \cline{2-8} 
\multicolumn{1}{|l|}{} &
  \multicolumn{1}{l|}{\textbf{Connectors}} &
  \multicolumn{1}{c|}{USB 3.0 Type-C} &
  \multicolumn{1}{c|}{\begin{tabular}[c]{@{}c@{}}USB 3.0 Micro-B \\ and SMA\end{tabular}} &
  \multicolumn{1}{c|}{USB 3.0 Micro-B} &
  \multicolumn{1}{c|}{M12 - 17 Pin} &
  \multicolumn{1}{c|}{USB 3.0 Type-C} &
  \multicolumn{1}{c|}{USB 3.0 Type-C} \\ \cline{2-8} 
\multicolumn{1}{|l|}{} &
  \multicolumn{1}{l|}{\textbf{Case Material}} &
  \multicolumn{1}{c|}{Aluminum alloy} &
  \multicolumn{1}{c|}{PCBs only} &
  \multicolumn{1}{c|}{PCBs only} &
  \multicolumn{1}{c|}{-} &
  \multicolumn{1}{c|}{Aluminum alloy} &
  \multicolumn{1}{c|}{Aluminum alloy} \\ \cline{2-8} 
\multicolumn{1}{|l|}{} &
  \multicolumn{1}{l|}{\textbf{Weight}} &
  \multicolumn{1}{c|}{40g} &
  \multicolumn{1}{c|}{} &
  \multicolumn{1}{c|}{} &
  \multicolumn{1}{c|}{180 g without lens} &
  \multicolumn{1}{c|}{102.6 g} &
  \multicolumn{1}{c|}{72.5 g} \\ \cline{2-8} 
\multicolumn{1}{|l|}{} &
  \multicolumn{1}{l|}{\textbf{Power Consumption}} &
  \multicolumn{1}{c|}{0.5W powered via USB} &
  \multicolumn{1}{c|}{powered via USB} &
  \multicolumn{1}{c|}{\begin{tabular}[c]{@{}c@{}}4.5W powered \\ via USB\end{tabular}} &
  \multicolumn{1}{c|}{-} &
  \multicolumn{1}{c|}{USB Power (VBUS)} &
  \multicolumn{1}{c|}{USB Power (VBUS)} \\ \cline{2-8} 
\multicolumn{1}{|l|}{} &
  \multicolumn{1}{l|}{\textbf{Sensor Tech}} &
  \multicolumn{1}{c|}{IMX636} &
  \multicolumn{1}{c|}{\begin{tabular}[c]{@{}c@{}}GenX320 CCAM5 module, \\ CM2 Flex Optical Module\end{tabular}} &
  \multicolumn{1}{c|}{IMX636} &
  \multicolumn{1}{c|}{\begin{tabular}[c]{@{}c@{}}PPS3MVCD \\ (Gen3.1 VGA sensor)\end{tabular}} &
  \multicolumn{1}{c|}{IMX636} &
  \multicolumn{1}{c|}{\begin{tabular}[c]{@{}c@{}}PPS3MVCD \\ (Gen3.1 VGA sensor)\end{tabular}} \\ \hline
\multicolumn{2}{|c|}{\multirow{3}{*}{\textbf{Other Features}}} &
  \multicolumn{3}{c|}{Synchronization interface} &
  \multicolumn{1}{c|}{-} &
  \multicolumn{1}{c|}{-} &
  \multicolumn{1}{c|}{-} \\ \cline{3-8} 
\multicolumn{2}{|c|}{} &
  \multicolumn{1}{c|}{-} &
  \multicolumn{1}{c|}{-} &
  \multicolumn{1}{c|}{-} &
  \multicolumn{1}{c|}{\begin{tabular}[c]{@{}c@{}}Programable with \\ dual-core ARM Cortex-A15 , \\ 1 × $\mu$SD Card $\geq$ 32 GB\end{tabular}} &
  \multicolumn{1}{c|}{-} &
  \multicolumn{1}{c|}{-} \\ \cline{3-8} 
\multicolumn{2}{|c|}{} &
  \multicolumn{6}{c|}{METAVISION Intelligence Suite - SDK support by PROPHESEE} \\ \hline \hline
\end{tabular}%
}
\end{table*}

Prophesee offers evaluation kits for exploring event-based vision, including USB cameras and embedded starter kits. USB cameras feature the Metavision EVK4-HD \cite{metavision_evK4_HD} with the IMX636 sensor (1280x720px) \cite{imx636_hd}, providing high dynamic range (>120 dB) and low pixel latency (<100 $\mu$s), the Metavision EVK3- GENX320  with the GenX320 sensor (320x320px) \cite{metavision_evk3_genx320}, known for ultra-low power consumption (down to 36 $\mu$W) and high dynamic range (>120 dB), and the Metavision EVK3-HD \cite{metavision_evk3_hd} with the IMX636 sensor and USB 3.0 connectivity. Embedded starter kits include the Metavision Starter Kit-AMD Kria KV260 \cite{metavision_kria_kv260}, combining IMX636 \cite{imx636} and GenX320 sensors for FPGA-based development, and the Metavision Starter Kit-STM32F7 \cite{metavision_stm32f7} , optimized for the STM32-F7 MCU with the GenX320 sensor for low-power applications. The Metavision SDK \cite{ metavision_sdk} offers a comprehensive suite of tools, including visualization applications, programming guides, and APIs in \texttt{C++} and Python for custom solution development and sample recordings. Tab. \ref{propheseeCams} summarizes the key characteristics and features of Prophesee’s event cameras.

\begin{figure}[t]
\centerline{\includegraphics[width=0.95\linewidth]{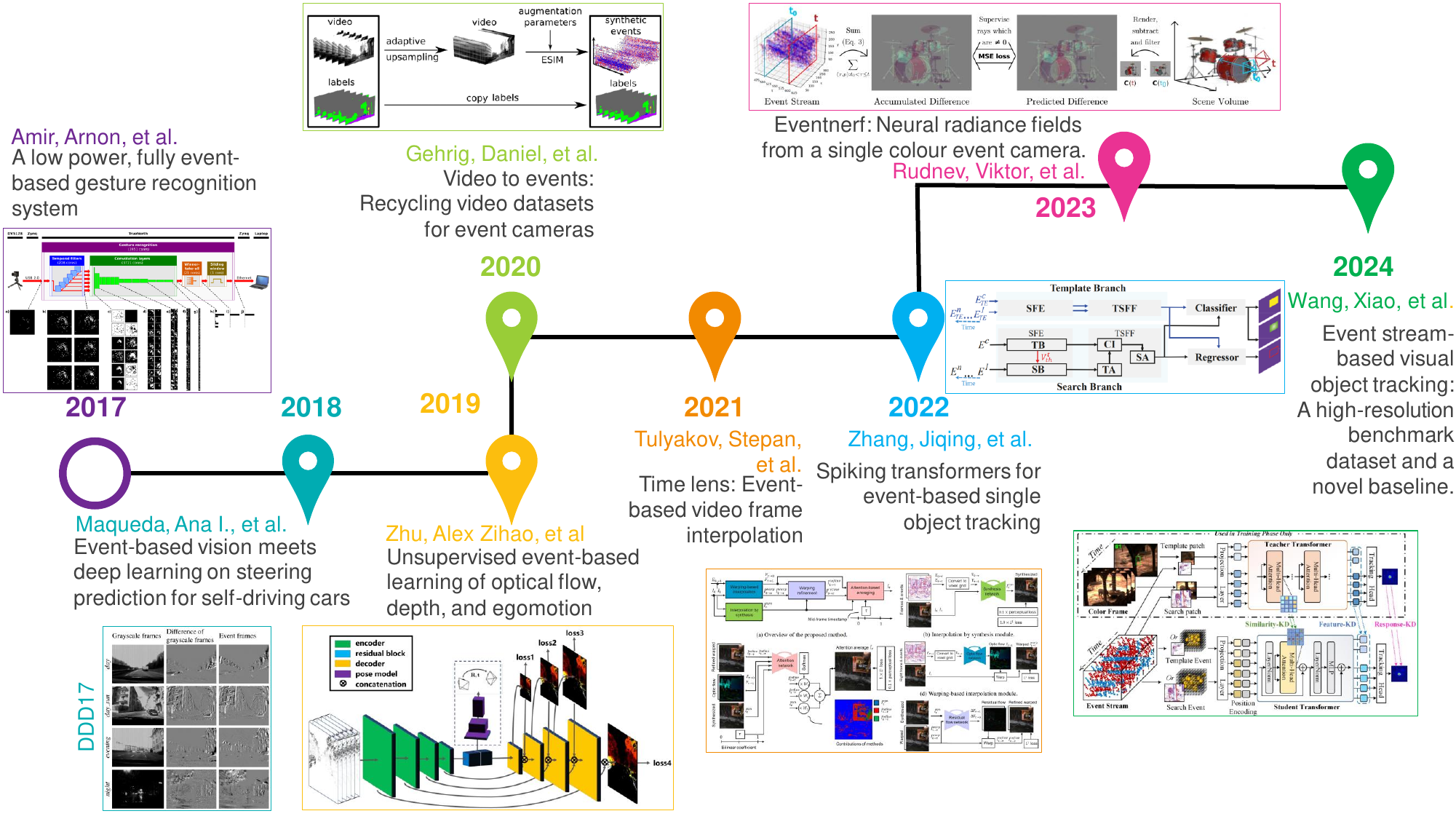}}
\caption{Key Milestone Papers and Works in Event-based Vision.}
\label{fig05}
\end{figure}

\section{Pioneering Progress: Milestones in Event-based Vision}
\label{eventMilestones}
In this section, significant milestone works in event-based vision from 2017 to 2024 (July) are reviewed, highlighting key advancements that have shaped the field, as illustrated in \cref{fig05}. In 2017, \cite{amir2017low} introduced a low-power, fully event-based gesture recognition system using the TrueNorth processor, achieving real-time accuracy with minimal power. \cite{mueggler2017event} released a comprehensive dataset and simulator, combining global-shutter and event-based sensors to advance algorithms for robotics and vision applications. \cite{li2017cifar10} developed the CIFAR10-DVS dataset, converting CIFAR-10 images into event streams, offering a valuable benchmark for event-driven object classification, utilizing a repeated closed-loop smooth (RCLS) movement of frame-based images.

In 2018, \cite{maqueda2018event} enhanced steering prediction for self-driving cars by adapting deep neural networks for event data. \cite{sironi2018hats} introduced HATS, a feature representation and machine learning architecture that improved object classification accuracy and released the first large real-world event-based dataset. \cite{zhu2018multivehicle} unveiled the multivehicle stereo event camera dataset (MVSEC), offering synchronized event streams and IMU data for $3D$ perception tasks. \cite{ rebecq2018esim}  developed ESIM, an open-source simulator for generating high-quality synthetic event data, while \cite{ zhu2018ev} also introduced EV-FlowNet, a self-supervised framework for optical flow estimation from event streams.
In 2019, \cite{zhu2019unsupervised} proposed an unsupervised learning framework for predicting optical flow and depth from event streams using discretized volume representation. \cite{rebecq2019events} developed a method to reconstruct high-quality videos from event data with a recurrent neural network for object classification and visual-inertial odometry. \cite{pan2019bringing}  introduced the event-based double integral (EDI) model to generate sharp, high-frame-rate videos from a single blurry frame and event data, addressing motion blur. Additionally, \cite{rebecq2019high} improved intensity image and color video reconstruction using a recurrent network trained on simulated data.

In 2020, \cite{ perot2020learning} released a high-resolution (1Mpx) dataset and a recurrent architecture with temporal consistency loss, improving object detection. \cite{ gehrig2020video} converted conventional video datasets into synthetic event data for detection and segmentation tasks, enhancing model training, while \cite{ scheerlinck2020fast}  developed a neural network for fast and efficient image reconstruction from event data. 
In 2021, \cite{gehrig2021dsec} introduced the high-resolution DSEC stereo dataset to improve autonomous driving under challenging lighting conditions. \cite{hu2021v2e} developed the v2e toolbox for generating realistic synthetic DVS events from intensity frames, enhancing object detection, particularly at night. \cite{tulyakov2021time}  proposed Time Lens, a frame interpolation method that improves image quality and handles dynamic scenarios. \cite{zhou2021event}  presented an event-based stereo-visual odometry system with real-time robustness. \cite{kim2021n} introduced the N-ImageNet dataset to support fine-grained object recognition with event cameras.

In 2022, \cite{zhang2022spiking} introduced STNet, a spiking transformer network for single-object tracking that combines global spatial and temporal cues for superior accuracy and speed. \cite{sun2022event} developed EFNet, a two-stage restoration network utilizing cross-modal attention, setting new benchmarks in motion deblurring with the REBlur dataset. \cite{schaefer2022aegnn} proposed AEGNN, which reduce computational complexity and latency by processing events as sparse, evolving spatiotemporal graphs. \cite{tulyakov2022time} presented Time Lens\texttt{++}, enhancing frame interpolation with parametric non-linear flow and multi-scale fusion.
In 2023, \cite{rudnev2023eventnerf} introduced EventNeRF, which uses a single-color event stream to achieve dense $3D$ reconstructions with high-quality RGB renderings. \cite{gehrig2023recurrent} developed recurrent vision transformers (RVT), reaching state-of-the-art object detection performance with reduced inference time and parameter efficiency. \cite{hwang2023ev} introduced Ev-NeRF, adapting neural radiance fields to event data for improved intensity image reconstruction under extreme conditions.

In 2024, \cite{wang2024event} introduced high-resolution data and hierarchical knowledge distillation to enhance speed and accuracy in visual object tracking. \cite{ aliminati2024sevd} (SEVD) provided synthetic multi-view data for robust traffic participant detection, while \cite{verma2024etram} (eTraM) offered 10 hr of event-based traffic monitoring data, showcasing the effectiveness of event cameras in diverse scenarios. These milestones demonstrate the rapid progress and growing potential of event-based vision technologies.

\begin{figure}[t]
\centerline{\includegraphics[width=0.68\linewidth]{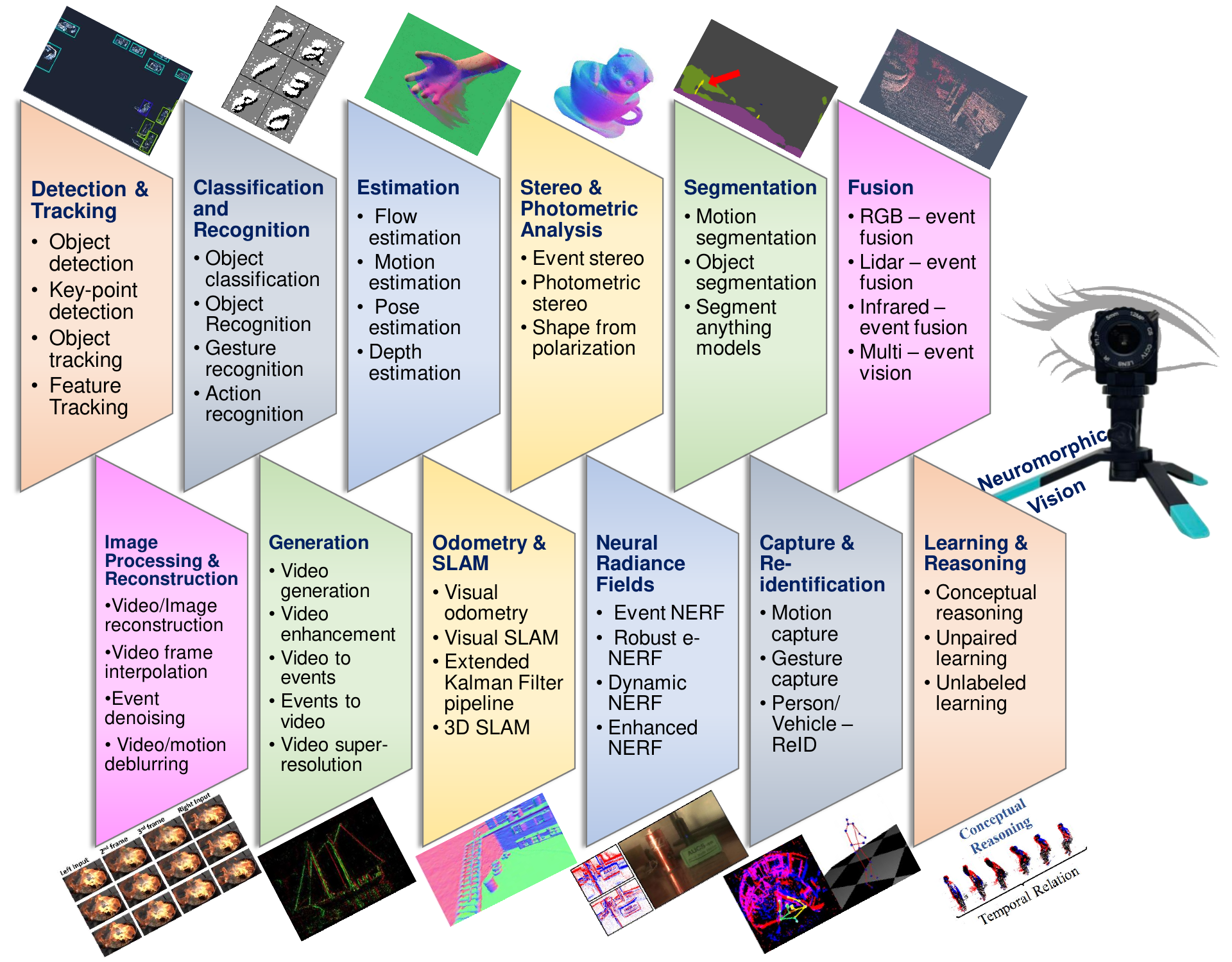}}
\caption{Showcasing Broad Applications and Notable Works in Event-based Vision Research.}
\label{fig06}
\end{figure}

\section{Event Cameras in Action: Diverse Tasks and Impacts}
\label{eventApplications}
Event-based vision is revolutionizing numerous fields by introducing new capabilities across a wide range of tasks, including detection, tracking, classification, recognition, and estimation. This section highlights key tasks, as illustrated in \cref{fig06}, and explores its significant impact across different application domains. In detection and tracking, event cameras with high temporal resolution and low latency have driven advancements in object detection, key-point detection, and tracking. Innovations such as scene-adaptive sparse transformers \cite{peng2024scene}, spiking \cite{zhang2022spiking} and recurrent vision transformers \cite{gehrig2023recurrent}, and self-supervised learning \cite{gao2024sd2event} have enhanced accuracy in these areas, benefiting applications in surveillance and autonomous driving \cite{chakravarthi2023event}. In classification and recognition, event cameras have markedly improved object classification, gesture and gait recognition, and action recognition, particularly in dynamic or complex scenes. Their ability to capture detailed temporal information boosts object classification through histograms of averaged time surfaces \cite{sironi2018hats} and space-time event clouds \cite{wang2019space}. 

Furthermore, event cameras greatly enhance estimation tasks such as optical flow, motion/pose, and depth estimation. The high-speed and low-latency characteristics of event cameras allow for precise calculation of motion, orientation, and depth, which are essential for understanding scene dynamics and improving $3D$ perception. Key advancements include progressive spatiotemporal alignment for motion estimation \cite{huang2023progressive}, globally optimal contrast maximization \cite{liu2020globally}, and tangentially elongated Gaussian belief propagation for optical flow \cite{10208126}. These developments are crucial for applications in robotics, augmented reality, and autonomous navigation. In stereo and photometric analysis, event-based vision supports advanced techniques like event stereo \cite{cho2023learning}, photometric stereo \cite{yu2024eventps}, and shape from polarization \cite{muglikar2023event}, offering high-resolution depth maps and detailed surface properties. For segmentation tasks, including semantic segmentation \cite{sun2022ess}, motion/object segmentation \cite{stoffregen2019event}, and segment anything models \cite{chen2024segment}, event cameras excel in dynamic and high-speed scenes, enabling precise scene understanding and object isolation. The fusion of event-based data with traditional frame-based \cite{xu2024dmr}, lidar, or infrared data \cite{geng2024event, zhou2024bring} further enhances applications such as environmental mapping by combining complementary information.

\begin{table}[t]
\caption{Event-based Vision Applications  with Related Notable Research Papers.}
\centering
\label{tasksTable}
\resizebox{0.95\textwidth}{!}{%
\begin{tabular}{c|c|l|l}
\toprule
 & 
  \textbf{Applications}  &
 \centering \textbf{Taks}  &
  \textbf{ Related Papers} \\ \midrule
  
\multirow{37}{*}{\rotatebox[origin=c]{90}{\centering \textbf{Event-based Vision}}} &

  \multirow{4}{*}{\begin{tabular}[c]{@{}c@{}}Detection \\ and \\ Tracking\end{tabular}} &
  Object detection & 
  \cite{peng2024scene, wu2024leod, gehrig2023recurrent, salvatore2022learned, verma2024etram, aliminati2024sevd, zubic2023chaos, perot2020learning, Zubic_2024_CVPR_SSM, GET_ICCV_2023, Schaefer22cvpr_aegnn, Gehrig24nature_DAGr, ASTM-NET, mesquida:cea-04321175_G2N2, 10160472, 10161164, 10161563, 9812059}\\ \cline{3-4} 
 &
   &
  Key-point detection &
  \cite{manderscheid2019speed, huang2023eventpoint, gao2024sd2event, keypoint_21_cvprw, eCDT_22_IROS, EDFLOW} \\ \cline{3-4} 
 &
   &
  Object tracking &
  \cite{wang2024event, zhang2022spiking, zhu2023cross, zhang2021object, zubic2023chaos, bagchi2020event, shah2024codedevents, 10161392} \\ \cline{3-4} 
 &
   &
  Feature tracking &
  \cite{li20243d, messikommer2023data, zihao2017event, seok2020robust, gehrig2018asynchronous, 10160768, 10161098} \\ \cline{2-4} 
 &
  \multirow{5}{*}{\begin{tabular}[c]{@{}c@{}}Classification\\ and\\ Recognition\end{tabular}} &
  Object classification &
  \cite{deng2022voxel,sironi2018hats}  
\hspace{80mm}                             
   \\ \cline{3-4} 
 &
   &
  Object recognition &
  \cite{cannici2019attention, zubic2023chaos, cho2023label, li2021graph,kim2021n, yao2021temporal, wan2022s2n, zheng2024eventdance, kim2022ev}\\ \cline{3-4} 
 &
   &
  Gesture recognition &
  \cite{amir2017low, wang2019space} \\ \cline{3-4} 
 &
   &
  Gait recognition &
  \cite{wang2019ev} \\ \cline{3-4} 
 &
   &
  Action recognition &
  \cite{zhou2024exact, plizzari2022e2, 9879993, 10480584} \\ \cline{2-4} 
 &
  \multirow{3}{*}{Estimation} &
  Optical flow estimation &
  \cite{pan2020single, gallego2019focus, gallego2018unifying, lee2020spike, ponghiran2023event, Luo_2024_CVPR, 10208126, 10376957, paredes2023taming, liu2023tma, luo2023learning, shiba2022secrets, 10160551, Shiba24pami} \\ \cline{3-4} 
 &
   &
  Motion\slash Pose estimation &
  \cite{gao2024n, huang2023progressive,liu2021spatiotemporal, liu2020globally, peng2020globally, zou2021eventhpe, rudneveventhands, nunes2023time, ren2024simple, elms2024event, 10160967, 10160531, 9812430, 9812142} \\ \cline{3-4} 
 &
   &
  Depth estimation &
  \cite{nam2022stereo, gallego2019focus, zhang2022data, cho2022selection, 10160605} \\ \cline{2-4} 
 &
  \multirow{3}{*}{\begin{tabular}[c]{@{}c@{}}Stereo and \\ Photometric\\ Analysis\end{tabular}} &
  Event stereo &
  \cite{cho2023learning, zhang2022discrete, nam2022stereo, andreopoulos2018low, wang2020stereo, zhou2018semi, tulyakov2019learning, mostafavi2021event, 9811687} \\ \cline{3-4} 
 &
   &
  Photometric stereo &
  \cite{yu2024eventps} \\ \cline{3-4} 
 &
   &
  Shape from Polarization &
  \cite{muglikar2023event, Mei_2023_CVPR} \\ \cline{2-4} 
 &
  \multirow{3}{*}{Segmentation} &
  Semantic segmentation &
  \cite{kong2024openess, jing2024hpl, sun2022ess, biswas2024halsie} \\ \cline{3-4} 
 &
  
   &
  Motion\slash Object segmentation &
  \cite{mitrokhin2020learning, stoffregen2019event, li2024event, zhou2021event} \\ \cline{3-4} 
 &
   &
  Segment Anything Models &
  \cite{chen2024segment}\\ \cline{2-4} 
 &
  \multirow{2}{*}{Fusion} &
  Frame\slash RGB – Event &
  \cite{xu2024dmr, zhang2023frame, tulyakov2022time, 10161098, 10161563, 9812059, 9523155} \\ \cline{3-4} 
 &
   &
  Lidar\slash Infrared  – Event &
  \cite{zhou2024bring, geng2024event, 10160681, 10161220, brebion2023learning, cocheteux2024muli}\\ \cline{2-4} 
 &
 
  \multirow{6}{*}{\begin{tabular}[c]{@{}c@{}}Reconstruction \\ and \\ Image \\ Processing\end{tabular}} 
  &
  
  Motion reconstruction &
  \cite{jiang2024complementing} \\ \cline{3-4} 
 &
   &
  Video reconstruction &
\cite{yang2023learning,wang2021asynchronous, weng2021event, rebecq2019events, zou2021learning, zhu2022event, liu2024seeing, 10462903} \\ \cline{3-4} 
 &
   &
  Image reconstruction\slash restoration &
  \cite{paredes2021back, wang2020eventsr, scheerlinck2020fast, nagata2020qr, Lou_2023_CVPR, liang2024towards, habuchi2024time, 9878722} \\ \cline{3-4} 
 &
   &
  Video frame interpolation &
  \cite{cho2024tta, liu2024video, kim2023event, tulyakov2022time, tulyakov2021time, wu2022video, yu2021training, yang2024latency, weng2023event, sun2023event, zhang2022unifying, he2022timereplayer, 10160276, 9523155} \\ \cline{3-4} 
 &
   &
  Event denoising &
  \cite{baldwin2020event, duan2024led, duan2021eventzoom, 9812003, 10138453, 9720086, 9893571} \\ \cline{3-4} 
 &
   &
  Motion\slash Video deblurring &
  \cite{zhang2022unifying, jiang2020learning, sun2022event, zhang2023generalizing, xu2021motion, kim2022event, cho2023non,  kim2024frequency, lin2020learning, EvShutter, EvUnroll} \\ \cline{2-4} 
 &
  \multirow{3}{*}{Generation} &
  Video generation\slash enhancement &
  \cite{wang2019event, liang2023coherent, wang2023unsupervised, 10160654} \\ \cline{3-4} 
 &
   &
  Video to events &
  \cite{gehrig2020video, zhang2023v2ce} \\ \cline{3-4} 
 &
  
   &
  Video\slash Event Super-resolution &
  \cite{lu2023learning, jing2021turning, han2021evintsr, li2021event, Han_2023_CVPR, huang2024bilateral} \\ \cline{2-4} 
 &
  \multirow{2}{*}{\begin{tabular}[c]{@{}c@{}}Odometry \\ and SLAM\end{tabular}} &
  Visual odometry &
  \cite{hidalgo2022event, liu2021spatiotemporal, 10160681, 9811805, 9811943, 10254473} \\ \cline{3-4} 
 &
   &
  SLAM &
  \cite{qu2024implicit, 9810191, Guo24tro, jiao2021comparing, 9809788} \\ \cline{2-4} 
 &
  NERF &
  Neural Radiance Fields &
  \cite{rudnev2023eventnerf, hwang2023ev, qi2023e2nerf, low2023robust, cannici2024mitigating, ma2023deformable} \\ \cline{2-4} 
 &
  \multirow{2}{*}{\begin{tabular}[c]{@{}c@{}}Capture and\\ Re-Identification\end{tabular}} &
  Motion capture &
 \cite{millerdurai2024eventego3d, xu2020eventcap} \\ \cline{3-4} 
 &
   &
  Person Re-ID &
  \cite{cao2023event, ahmad2023person} 
  \\

  \cline{2-4} 
 &
  \multirow{3}{*}{\begin{tabular}[c]{@{}c@{}}Learning and \\ Reasoning\end{tabular}} &
  Conceptual reasoning &
  \cite{zhou2024exact} \\ \cline{3-4} 
 &
   &
  Unsupervised learning &
  \cite{zhu2019unsupervised, gallego2019focus, klenk2024masked} \\ 
  
  \cline{3-4} &
   &
  Unlabeled and unpaired event data
  &
\cite{wang2021evdistill}
   \\ \bottomrule
\end{tabular}%
}
\end{table}

Event-based vision significantly advances reconstruction and image processing tasks, contributing to video reconstruction \cite{zhu2022event, weng2021event}, image reconstruction \cite{paredes2021back, wang2020eventsr}, video frame interpolation \cite{yu2021training, liu2024video}, event denoising \cite{baldwin2020event}, and motion deblurring \cite{sun2022event, cho2023non}. In generation-related tasks, it aids in video generation and enhancement \cite{wang2019event, liu2017high}, video-to-events conversion \cite{gehrig2020video}, and super-resolution \cite{huang2024bilateral, lu2023learning}, facilitating high-quality content creation and analysis. For odometry and SLAM, event-based vision plays a crucial role in visual odometry \cite{9811805} and simultaneous localization and mapping \cite{9810191}, providing accurate navigation and mapping capabilities. 
\cref{tasksTable} highlights notable works using event cameras across various tasks and application domains, underscoring the transformative impact of event-based vision in addressing complex challenges and driving innovation.

\begin{table}[t]
\centering
\caption{Summary of Real-world Event-based Datasets.}
\label{realWorldEventData}
\resizebox{0.95\textwidth}{!}{%
\begin{tabular}{lllllll}
\toprule
\multicolumn{1}{c|}{\textbf{Year}} &
  \multicolumn{1}{c|}{\textbf{\begin{tabular}[c]{@{}c@{}}Dataset \\ Name\end{tabular}}} &
  \multicolumn{1}{c|}{\textbf{\begin{tabular}[c]{@{}c@{}}Event Modality \\ Used\end{tabular}}} &
  \multicolumn{1}{c|}{\textbf{\begin{tabular}[c]{@{}c@{}}Subject/Object \\ Classes\end{tabular}}} &
  \multicolumn{1}{c|}{\textbf{Labeling}} &
  \multicolumn{1}{c|}{\textbf{Tasks}} &
  \multicolumn{1}{c}{\textbf{\begin{tabular}[c]{@{}c@{}}Dataset \\ Description\end{tabular}}} \\ \midrule
\multicolumn{1}{l|}{2024} &
  \multicolumn{1}{l|}{\begin{tabular}[c]{@{}l@{}}EventVOT  {\cite{wang2024event}}\end{tabular}} &
  \multicolumn{1}{l|}{\begin{tabular}[c]{@{}l@{}}Prophesee EVK4\\ HD\end{tabular}} &
  \multicolumn{1}{l|}{\begin{tabular}[c]{@{}l@{}}UAV’s, Pedestrians, Vehicles, \\ Ball sports\end{tabular}} &
  \multicolumn{1}{l|}{Yes, Manual} &
  \multicolumn{1}{l|}{Object tracking} &
  \begin{tabular}[c]{@{}l@{}}Large-scale, high-resolution, visual object-tracking \\ dataset\end{tabular} \\ \hline
\multicolumn{1}{l|}{2024} &
  \multicolumn{1}{l|}{eTraM {\cite{verma2024etram}}} &
  \multicolumn{1}{l|}{\begin{tabular}[c]{@{}l@{}}Prophesee EVK4\\ HD\end{tabular}} &
  \multicolumn{1}{l|}{Vehicles, Pedestrians, Micro-mobility} &
  \multicolumn{1}{l|}{\begin{tabular}[c]{@{}l@{}}Yes, Manual, \\ $2D$ BB\end{tabular}} &
  \multicolumn{1}{l|}{\begin{tabular}[c]{@{}l@{}}Object detection, \\ tracking\end{tabular}} &
  \begin{tabular}[c]{@{}l@{}}High-resolution, large-scale, fixed   Traffic perception \\ dataset for traffic monitoring\end{tabular} \\ \hline
\multicolumn{1}{l|}{2024} &
  \multicolumn{1}{l|}{SeAct {\cite{zhou2024exact}}} &
  \multicolumn{1}{l|}{DAVIS346} &
  \multicolumn{1}{l|}{\begin{tabular}[c]{@{}l@{}}Human actions like sit, catch, throw, \\ vomit, handshake\end{tabular}} &
  \multicolumn{1}{l|}{Yes} &
  \multicolumn{1}{l|}{Action recognition} &
  Event-text action recognition dataset \\ \hline
\multicolumn{1}{l|}{2023} &
  \multicolumn{1}{l|}{PEDRo {\cite{boretti2023pedro}}} &
  \multicolumn{1}{l|}{DAVIS346} &
  \multicolumn{1}{l|}{Persons} &
  \multicolumn{1}{l|}{\begin{tabular}[c]{@{}l@{}}Yes, Manual, \\ $2D$ BB\end{tabular}} &
  \multicolumn{1}{l|}{Objected detection} &
  \begin{tabular}[c]{@{}l@{}}A large person detection dataset recorded with a moving\\  camera\end{tabular} \\ \hline
  
\multicolumn{1}{l|}{2022} &
  \multicolumn{1}{l|}{\begin{tabular}[c]{@{}l@{}}DVS-Lip {\cite{tan2022multi}}\end{tabular}} &
  \multicolumn{1}{l|}{DAVIS346} &
  \multicolumn{1}{l|}{\begin{tabular}[c]{@{}l@{}}Different sequences containing all \\ words in English vocabulary\end{tabular}} &
  \multicolumn{1}{l|}{Yes} &
  \multicolumn{1}{l|}{Gesture recognition} &
  \begin{tabular}[c]{@{}l@{}}Lip-reading dataset with over 100 classes, 40 speakers, \\ 19K plus utterances in the English language\end{tabular} \\ \hline
  
\multicolumn{1}{l|}{2022} &
  \multicolumn{1}{l|}{EVIMO2{\cite{burner2022evimo2}}} &
  \multicolumn{1}{l|}{\begin{tabular}[c]{@{}l@{}}Prophesee Gen3\\ DVS Gen3\end{tabular}} &
  \multicolumn{1}{l|}{Moving objects} &
  \multicolumn{1}{l|}{\begin{tabular}[c]{@{}l@{}}Yes \\ Automatic\end{tabular}} &
  \multicolumn{1}{l|}{\begin{tabular}[c]{@{}l@{}}Motion Segmentation, \\ Recognition\end{tabular}} &
  \begin{tabular}[c]{@{}l@{}}Indoor dataset with moving objects, 3D motion of the  \\ sensor, object motion and structure, masks and optical flow \end{tabular} \\ \hline
  
\multicolumn{1}{l|}{2021} &
  \multicolumn{1}{l|}{DSEC {\cite{gehrig2021dsec}}} &
  \multicolumn{1}{l|}{\begin{tabular}[c]{@{}l@{}}Prophesee Gen \\ 3.1\end{tabular}} &
  \multicolumn{1}{l|}{\begin{tabular}[c]{@{}l@{}}Driving scenarios with diverse \\ illumination during day \& nighttime\end{tabular}} &
  \multicolumn{1}{l|}{Yes} &
  \multicolumn{1}{l|}{Stereo matching} &
  High-resolution stereo dataset for driving scenarios \\ \hline
\multicolumn{1}{l|}{2020} &
  \multicolumn{1}{l|}{GEN1 {\cite{de2020large}}} &
  \multicolumn{1}{l|}{\begin{tabular}[c]{@{}l@{}}Prophesee \\ Gen 1\end{tabular}} &
  \multicolumn{1}{l|}{Cars, Pedestrians} &
  \multicolumn{1}{l|}{\begin{tabular}[c]{@{}l@{}}Yes, Manual,\\ $2D$ BB\end{tabular}} &
  \multicolumn{1}{l|}{\begin{tabular}[c]{@{}l@{}}Object detection, tracking, \\ flow estimation\end{tabular}} &
  \begin{tabular}[c]{@{}l@{}}Large event-based automotive (cars and pedestrians) \\ detection dataset\end{tabular} \\ \hline
\multicolumn{1}{l|}{2020} &
  \multicolumn{1}{l|}{1 Mpx {\cite{perot2020learning}}} &
  \multicolumn{1}{l|}{\begin{tabular}[c]{@{}l@{}}Prophesee \\ 1 Mpx\end{tabular}} &
  \multicolumn{1}{l|}{\begin{tabular}[c]{@{}l@{}}Pedestrians, two-wheelers, cars, \\ trucks, buses, traffic signs, traffic lights\end{tabular}} &
  \multicolumn{1}{l|}{\begin{tabular}[c]{@{}l@{}}Yes, Automatic,\\ $2D$ BB\end{tabular}} &
  \multicolumn{1}{l|}{Object detection} &
  \begin{tabular}[c]{@{}l@{}}High-resolution large-scale automotive detection \\ dataset (25M BB, 14 hr)\end{tabular} \\ \hline
\multicolumn{1}{l|}{2019} &
  \multicolumn{1}{l|}{\begin{tabular}[c]{@{}l@{}}ASL-DVS {\cite{bi2019graph}}\end{tabular}} &
  \multicolumn{1}{l|}{DAVIS240c} &
  \multicolumn{1}{l|}{\begin{tabular}[c]{@{}l@{}}24 classes of hand gesture correspond \\ to 24 letters (A-Y, excluding J)\end{tabular}} &
  \multicolumn{1}{l|}{Yes} &
  \multicolumn{1}{l|}{Gesture recognition} &
  \begin{tabular}[c]{@{}l@{}}Dataset for American Sign Language (ASL) \\ recognition\end{tabular} \\ \hline

  \multicolumn{1}{l|}{2019} &
  \multicolumn{1}{l|}{EV-IMO{\cite{mitrokhin2019ev}}} &
  \multicolumn{1}{l|}{\begin{tabular}[c]{@{}l@{}}DAVIS346\\ \end{tabular}} &
  \multicolumn{1}{l|}{Moving objects} &
  \multicolumn{1}{l|}{\begin{tabular}[c]{@{}l@{}}Yes \\ Automatic\end{tabular}} &
  \multicolumn{1}{l|}{\begin{tabular}[c]{@{}l@{}}Motion segmentation,  visual\\  odometry, optical flow, stereo\end{tabular}} &
  \begin{tabular}[c]{@{}l@{}}Indoor dataset with moving objects, 3D motion of the   \\ sensor, 3D object and structure, and object masks \end{tabular} \\ \hline

\multicolumn{1}{l|}{2018} &
  \multicolumn{1}{l|}{\begin{tabular}[c]{@{}l@{}}N-Cars {\cite{sironi2018hats}}\end{tabular}} &
  \multicolumn{1}{l|}{\begin{tabular}[c]{@{}l@{}}Prophesee \\ Gen 1\end{tabular}} &
  \multicolumn{1}{l|}{Cars} &
  \multicolumn{1}{l|}{\begin{tabular}[c]{@{}l@{}}Yes, $2D$ BB,\\ Semi-automatic\end{tabular}} &
  \multicolumn{1}{l|}{Object Classification} &
  \begin{tabular}[c]{@{}l@{}}Large real-world event-based dataset for object \\ classification\end{tabular} \\ \hline
\multicolumn{1}{l|}{2018} &
  \multicolumn{1}{l|}{\begin{tabular}[c]{@{}l@{}}MVSEC  {\cite{zhu2018multivehicle}}\end{tabular}} &
  \multicolumn{1}{l|}{DAVIS346B} &
  \multicolumn{1}{l|}{Poses and depth} &
  \multicolumn{1}{l|}{Yes} &
  \multicolumn{1}{l|}{\begin{tabular}[c]{@{}l@{}}Feature tracking, visual \\ odometry, depth estimation\end{tabular}} &
  \begin{tabular}[c]{@{}l@{}}Synchronized stereo pair dataset recorded\\ from a handheld rig, hexacopter, top of a car, on a \\ motorcycle, in different illumination levels\end{tabular} \\ \hline
\multicolumn{1}{l|}{2017} &
  \multicolumn{1}{l|}{\begin{tabular}[c]{@{}l@{}}DDD17 {\cite{binas2017ddd17}}\end{tabular}} &
  \multicolumn{1}{l|}{DAVIS346B} &
  \multicolumn{1}{l|}{Vehicle speed, driver steering etc.,} &
  \multicolumn{1}{l|}{Yes} &
  \multicolumn{1}{l|}{Steering angle prediction} &
  \begin{tabular}[c]{@{}l@{}}driving recordings in highways, city during day, \\ evening, night, dry, and wet weather conditions\end{tabular} \\ \hline
\multicolumn{1}{l|}{2017} &
  \multicolumn{1}{l|}{\begin{tabular}[c]{@{}l@{}}DvsGesture{\cite{amir2017low}}\end{tabular}} &
  \multicolumn{1}{l|}{DVS128} &
  \multicolumn{1}{l|}{11 hand and arm gesture classes} &
  \multicolumn{1}{l|}{Yes} &
  \multicolumn{1}{l|}{Gesture recognition} &
  \begin{tabular}[c]{@{}l@{}}Gesture dataset comprising 11 hand gesture categories \\ from 29 subjects under 3 illumination conditions\end{tabular} \\ \hline
\multicolumn{1}{l|}{2017} &
  \multicolumn{1}{l|}{\begin{tabular}[c]{@{}l@{}}Event Camera \\ Dataset {\cite{mueggler2017event}}\end{tabular}} &
  \multicolumn{1}{l|}{DAVIS 240C} &
  \multicolumn{1}{l|}{\begin{tabular}[c]{@{}l@{}}Object rotation, translation, person\\ walking and running, etc.,\end{tabular}} &
  \multicolumn{1}{l|}{Yes} &
  \multicolumn{1}{l|}{\begin{tabular}[c]{@{}l@{}}Pose estimation, visual\\ odometry, SLAM\end{tabular}} &
  \begin{tabular}[c]{@{}l@{}}The object's motion captured in outdoor and indoor \\ scenarios with varying speed and DoFs\end{tabular} \\ \bottomrule
\end{tabular}%
}
\end{table}

\section{Data for Innovation: Event-based Vision Datasets}
\label{eventDatasets}
Event-based vision datasets are crucial for advancing the field by providing resources for training and evaluating algorithms. Real-world datasets, captured with event cameras, cover diverse scenarios, while synthetic datasets from simulators offer controlled data for experimentation. This section reviews prominent datasets, summarized in Tab. \ref{realWorldEventData} and Tab. \ref{synEventDataset}, with a detailed list available on the GitHub page.

\subsection{Real-world Datasets}
The EventVOT \cite {wang2024event} dataset offers high-resolution visual object tracking data using the Prophesee EVK4 HD camera, covering diverse target categories such as UAVs, pedestrians, vehicles, and ball sports across various motion speeds and lighting conditions. eTraM \cite{verma2024etram} dataset provides a comprehensive traffic monitoring dataset with 10 hr of data from the Prophesee EVK4 HD camera, including $2M$ bounding box annotations across eight traffic participant classes. SeAct \cite{ zhou2024exact} introduces a semantic-rich dataset for event-text action recognition, collected with a DAVIS 346 camera and enhanced with GPT-4 generated action captions. 
DVS-Lip \cite{ tan2022multi} is a lip-reading dataset recorded with the DAVIS 346 camera, featuring $100$ words and fine-grained movement information. DSEC \cite{gehrig2021dsec} provides stereo data for driving scenarios, including lidar and GPS measurements, with $53$ sequences collected under various illumination conditions. GEN1 \cite{de2020large} offers a large-scale automotive detection dataset with over $39$ hr of data collected in different driving conditions. 

The 1 MPX \cite{ perot2020learning} dataset includes high-resolution data from a 1-megapixel event camera, providing $25M$ bounding boxes for object detection in automotive scenarios. N-Cars \cite{sironi2018hats} features recordings of urban environments for object classification, capturing $80$ min of video with an ATIS camera. MVSEC \cite{zhu2018multivehicle} includes synchronized stereo data for $3D$ perception across varied environments, while DDD17 \cite{binas2017ddd17} provides both event and frame-based driving data with over $12$ hr of recordings. DvsGesture \cite{amir2017low} is a gesture recognition dataset with $1,342$ instances of $11$ hand and arm gestures recorded under different lighting conditions using the DVS 128 camera. Moreover, the Event Camera Dataset \cite{mueggler2017event} offers data for pose estimation, visual odometry, and SLAM using a DAVIS camera.

\begin{table}[t]
\caption{Summary of Synthetic Event-based Datasets.}
\centering
\label{synEventDataset}
\resizebox{0.95\textwidth}{!}{%
\begin{tabular}{lllllll}
\toprule
\multicolumn{1}{c|}{\textbf{Year}} &
  \multicolumn{1}{c|}{\textbf{\begin{tabular}[c]{@{}c@{}}Dataset\\ Name\end{tabular}}} &
  \multicolumn{1}{c|}{\textbf{\begin{tabular}[c]{@{}c@{}}Event Modality\\ Used\end{tabular}}} &
  \multicolumn{1}{c|}{\textbf{\begin{tabular}[c]{@{}c@{}}Subject/Object\\ Classes\end{tabular}}} &
  \multicolumn{1}{c|}{\textbf{Labeling}} &
  \multicolumn{1}{c|}{\textbf{Task}} &
  \multicolumn{1}{c}{\textbf{Dataset Description}} \\ \midrule
\multicolumn{1}{l|}{2024} &
  \multicolumn{1}{l|}{SEVD \cite{aliminati2024sevd}} &
  \multicolumn{1}{l|}{CARLA DVS} &
  \multicolumn{1}{l|}{\begin{tabular}[c]{@{}l@{}}Car, truck, van, bicycle, motorcycle, \\ pedestrian\end{tabular}} &
  \multicolumn{1}{l|}{\begin{tabular}[c]{@{}l@{}}Yes, Automatic labeling, \\ $2D$ and $3D$ BB\end{tabular}} &
  \multicolumn{1}{l|}{\begin{tabular}[c]{@{}l@{}}Object detection, \\ tracking\end{tabular}} &
  Multi-view ego and fixed perception dataset for traffic monitoring \\ \hline
\multicolumn{1}{l|}{2024} &
  \multicolumn{1}{l|}{Event-KITTI \cite{zhou2024bring}} &
  \multicolumn{1}{l|}{V2E} &
  \multicolumn{1}{l|}{Vehicles, pedestrians, cyclists, etc.,} &
  \multicolumn{1}{l|}{Yes} &
  \multicolumn{1}{l|}{\begin{tabular}[c]{@{}l@{}}Object Detection, \\ tracking\end{tabular}} &
  \begin{tabular}[c]{@{}l@{}}An event-based version of KITTI using the V2E for daytime images \\ and a noise model for nighttime images of corresponding daytime\end{tabular} \\ \hline
\multicolumn{1}{l|}{2023} &
\multicolumn{1}{l|}{\begin{tabular}[c]{@{}l@{}}ESfP-Synthetic\\{\cite{muglikar2023event}}\end{tabular}}
   &
  \multicolumn{1}{l|}{ESIM} &
  \multicolumn{1}{l|}{\begin{tabular}[c]{@{}l@{}}Scenes consisting of a textured mesh \\ illuminated with a point light source\end{tabular}} &
  \multicolumn{1}{l|}{\begin{tabular}[c]{@{}l@{}}Yes, ground-truth surface \\ normal from renderer\end{tabular}} &
  \multicolumn{1}{l|}{Object reconstruction} &
  \begin{tabular}[c]{@{}l@{}}Dataset for Surface normal estimation using event-based shape from \\ polarization\end{tabular} \\ \hline
\multicolumn{1}{l|}{2022} &
  \multicolumn{1}{l|}{\begin{tabular}[c]{@{}l@{}}N-EPIC \\ Kitchen {\cite{plizzari2022e2}}\end{tabular}}  &
  \multicolumn{1}{l|}{ESIM} &
  \multicolumn{1}{l|}{\begin{tabular}[c]{@{}l@{}}8 action classes (Put, take, open, close, \\ wash, cut, mix, pour)\end{tabular}} &
  \multicolumn{1}{l|}{Yes} &
  \multicolumn{1}{l|}{Action Recognition} &
  \begin{tabular}[c]{@{}l@{}}Event-based camera extension of the large-scale EPIC-Kitchens \\ dataset\end{tabular} \\ \hline
\multicolumn{1}{l|}{2021} &
  \multicolumn{1}{l|}{N-ImageNet \cite{kim2021n}} &
  \multicolumn{1}{l|}{\begin{tabular}[c]{@{}l@{}}Samsung DVS \\ Gen3\end{tabular}} &
  \multicolumn{1}{l|}{1000 Object Classes (same as ImageNet)} &
  \multicolumn{1}{l|}{Yes} &
  \multicolumn{1}{l|}{Object Recognition} &
  Event-based version of the original ImageNet dataset. \\ \hline
\multicolumn{1}{l|}{2017} &
\multicolumn{1}{l|}{\begin{tabular}[c]{@{}l@{}}CIFAR10-DVS\\{\cite{li2017cifar10}}\end{tabular}}  &
  \multicolumn{1}{l|}{DVS camera} &
  \multicolumn{1}{l|}{\begin{tabular}[c]{@{}l@{}}10 Object Classes (airplane, automobile, \\ bird, cat, deer, dog, frog, horse, ship, truck)\end{tabular}} &
  \multicolumn{1}{l|}{Yes} &
  \multicolumn{1}{l|}{Object Classification} &
  Event-based representations of the original CIFAR-10 images \\ \hline
\multicolumn{1}{l|}{2015} &
  \multicolumn{1}{l|}{N-MNIST \cite{orchard2015converting}} &
  \multicolumn{1}{l|}{DVS camera} &
  \multicolumn{1}{l|}{10 classes of digits (0-9)} &
  \multicolumn{1}{l|}{Yes} &
  \multicolumn{1}{l|}{Object Classification} &
  An event-based version of the original MNIST dataset \\ \hline
\multicolumn{1}{l|}{2015} &
  \multicolumn{1}{l|}{N-Caltech 101 \cite{orchard2015converting}} &
  \multicolumn{1}{l|}{\begin{tabular}[c]{@{}l@{}}ATIS image \\ sensor\end{tabular}} &
  \multicolumn{1}{l|}{101 Object Classes (animals, vehicles, etc.,)} &
  \multicolumn{1}{l|}{Yes} &
  \multicolumn{1}{l|}{Object Classification} &
  Event-based version of the traditional Caltech101 dataset \\ \hline
\multicolumn{1}{l|}{2015} &
  \multicolumn{1}{l|}{MNIST-DVS \cite{serrano2015poker}} &
  \multicolumn{1}{l|}{DVS camera} &
  \multicolumn{1}{l|}{10 classes of digits (0-9)} &
  \multicolumn{1}{l|}{Yes} &
  \multicolumn{1}{l|}{Object Classification} &
  An event-based version of the original MNIST dataset \\ \bottomrule
\end{tabular}%
}
\end{table}

\subsection{Synthetic Datasets}
The SEVD \cite{aliminati2024sevd} dataset provides a comprehensive synthetic event-based vision dataset using multiple DVS cameras within the CARLA simulator. It captures multi-view data across various lighting and weather conditions for ego and fixed traffic perception, including RGB imagery, depth maps, optical flow, and segmentation annotations to facilitate diverse traffic monitoring. The Event-KITTI \cite{zhou2024bring} dataset extends the KITTI by generating event streams from daytime and synthesizing nighttime images, aiding in scene flow analysis and motion fusion. The ESfP-Synthetic \cite{ muglikar2023event} dataset focuses on shape from polarization by rendering scenes with a polarizer and using ESIM to simulate events.

The N-ImageNet \cite{kim2021n} dataset, derived from ImageNet using a moving event camera setup, serves as a benchmark for fine-grained object recognition, addressing artifacts from monitor refresh mechanisms. The CIFAR10-DVS \cite{li2017cifar10} dataset converts CIFAR-10 into event streams, offering an intermediate difficulty dataset for event-driven object classification through realistic image movements. Lastly, the N-MNIST and N-Caltech \cite{orchard2015converting} datasets convert MNIST and Caltech101 into spiking neuromorphic datasets using a pan-tilt camera platform, facilitating studies on neuromorphic vision and sensor motion. These synthetic datasets have collectively advanced event-based vision, supporting diverse applications.

\begin{table}[h]
\caption{Summary of Open-source Event Camera Simulators.}
\centering
\label{eventSimulators_table}
\resizebox{0.95\textwidth}{!}{%
\begin{tabular}{l|l|l|l|l|l|l|l}
\toprule
\multicolumn{1}{c|}{\textbf{Simulator}} &
  \multicolumn{1}{c|}{\textbf{\begin{tabular}[c]{@{}c@{}}Open \\ Source\end{tabular}}} &
  \multicolumn{1}{c|}{\textbf{\begin{tabular}[c]{@{}c@{}}Programming \\ Language\end{tabular}}} &
  \multicolumn{1}{c|}{\textbf{Input}} &
  \multicolumn{1}{c|}{\textbf{Output}} &
  \multicolumn{1}{c|}{\textbf{Dependencies}} &
  \multicolumn{1}{c|}{\textbf{Description}} &
  \multicolumn{1}{c}{\textbf{\begin{tabular}[c]{@{}c@{}}Related \\ Resources\end{tabular}}} \\ \midrule
\textbf{ESIM} &
  Yes &
  \texttt{C++} &
  \begin{tabular}[c]{@{}l@{}}Arbitrary $3D$ camera motion, \\ initial images\end{tabular} &
  \begin{tabular}[c]{@{}l@{}}Event streams, standard images, \\ IMU data, ground truth\end{tabular} &
  \begin{tabular}[c]{@{}l@{}}OpenGL Renderer / \\ UnrealCVRenderer\end{tabular} &
  \begin{tabular}[c]{@{}l@{}}Simulates arbitrary camera motion in $3D$ scenes, while providing \\ events, standard images, inertial measurements, with ground-truth\end{tabular} &
\cite{rebecq2018esim, esim2023} \\ \hline
\textbf{\begin{tabular}[c]{@{}l@{}}DAVIS \\ Simulator\end{tabular}} &
  Yes &
  \texttt{Python} &
  \begin{tabular}[c]{@{}l@{}}Camera trajectory, Blender \\ scenes, render configurations\end{tabular} &
  \begin{tabular}[c]{@{}l@{}}Event streams, camera calibration, \\ ground truth, intensity images, depth map\end{tabular} &
  Blender &
  \begin{tabular}[c]{@{}l@{}}Generates synthetic DVS datasets using Blender for prototyping \\ visual odometry or event-based feature tracking algorithms\end{tabular} &
 \cite{mueggler2017event, davis_simulator2023} \\ \hline
\textbf{v2e} &
  Yes &
  \texttt{Python} &
  \begin{tabular}[c]{@{}l@{}}RGB or grayscale videos, \\ image sequences\end{tabular} &
  event frame videos (.h5 format) &
  \begin{tabular}[c]{@{}l@{}}PyTorch, OpenCV,\\ \texttt{Python} packages\end{tabular} &
  \begin{tabular}[c]{@{}l@{}}synthesizes  event data from any real (or synthetic) conventional \\ frame-based video using pre-trained DVS pixel model\end{tabular} &
  \cite{hu2021v2e, v2e2023} \\ \hline
\textbf{\begin{tabular}[c]{@{}l@{}}ICNS \\ Simulator\end{tabular}} &
  Yes &
  \begin{tabular}[c]{@{}l@{}}\texttt{Python}\slash \\ \texttt{C++}\slash Matlab\end{tabular} &
  \begin{tabular}[c]{@{}l@{}}Videos, Blender-generated \\ scenes, camera trajectories\end{tabular} &
  \begin{tabular}[c]{@{}l@{}}Event streams, camera calibration, \\ ground truth, intensity images\end{tabular} &
  Blender &
  \begin{tabular}[c]{@{}l@{}}an extended DVS pixel simulator for neuromorphic benchmarks \\ which simplifies the latency and the noise models\end{tabular} &
\cite{joubert2021event, iebcs2023}  \\ \hline
\textbf{\begin{tabular}[c]{@{}l@{}}V2CE \\ Toolbox\end{tabular}} &
  Yes &
  \texttt{Python} &
  \begin{tabular}[c]{@{}l@{}}RGB or gray-scale videos, \\ image sequences\end{tabular} &
  Event streams &
  PyTorch &
  \begin{tabular}[c]{@{}l@{}}Converts RGB or gray-scale videos to event streams. Trained on \\ MVSEC dataset, optimized for DAVIS 346B cameras (346x260px).\end{tabular} &
  \cite{zhang2023v2ce, v2ce2023}  \\ \hline
\textbf{\begin{tabular}[c]{@{}l@{}}DVS-\\ Voltmeter\end{tabular}} &
  Yes &
  \texttt{Python} &
  High frame-rate videos &
  Event streams &
  PyTorch, OpenCV &
  \begin{tabular}[c]{@{}l@{}}Stochastic DVS simulator, incorporating voltage variations, \\ randomness from photon reception, and noise effects.\end{tabular} &
  \cite{lin2022dvs, dvs_voltmeter2023} \\ \hline
\textbf{\begin{tabular}[c]{@{}l@{}}Carla: \\ DVS Camera\end{tabular}} &
  Yes &
  \begin{tabular}[c]{@{}l@{}}\texttt{Python}\slash  \\ \texttt{C++}\end{tabular} &
  \begin{tabular}[c]{@{}l@{}}Synchronous frames from \\ a video within CARLA\end{tabular} &
  Event streams (carla.DVSEventArray) &
  \begin{tabular}[c]{@{}l@{}}CARLA simulator,\\ \texttt{Python} packages\end{tabular} &
  \begin{tabular}[c]{@{}l@{}}Emulates a DVS intensity changes asynchronously, providing \\ microsecond temporal resolution.\end{tabular} &
  \cite{carla_dvs2023} \\ \hline
\textbf{\begin{tabular}[c]{@{}l@{}}Prophesee: \\ VtoE\end{tabular}} &
  Yes &
  \texttt{Python} &
  \begin{tabular}[c]{@{}l@{}}Image (png/jpg) or video \\ (mp4/avi)\end{tabular} &
  Event-based frame\slash video &
  \begin{tabular}[c]{@{}l@{}}Metavision SDK, \\ \texttt{Python} packages\end{tabular} &
  \begin{tabular}[c]{@{}l@{}}Transforms frame-based images or videos into event-based \\ counterpart\end{tabular} &
\cite{prophesee_event_simulator2023} \\ \hline
  \textbf{\begin{tabular}[c]{@{}l@{}}DVS-\\ Voltmeter\end{tabular}} &
  Yes &
  \texttt{Python} &
  High frame-rate videos &
  Event streams &
  PyTorch, OpenCV &
  \begin{tabular}[c]{@{}l@{}}Stochastic DVS simulator, incorporating voltage variations, \\ randomness from photon reception, and noise effects.\end{tabular} &
  \cite{lin2022dvs, dvs_voltmeter2023} \\ \hline
  
\textbf{\begin{tabular}[c]{@{}l@{}}VISTA\end{tabular}} &
  Yes &
  \begin{tabular}[c]{@{}l@{}}\texttt{Python}\slash  \\ \texttt{C++}\end{tabular} &
  \begin{tabular}[c]{@{}l@{}}Synchronous frames from \\ a video within VISTA\end{tabular} &
  Event streams &
  \begin{tabular}[c]{@{}l@{}}VISTA simulator,\\ \texttt{Python} packages\end{tabular} &
  \begin{tabular}[c]{@{}l@{}}Synthesizes event data locally around RGB data given a viewpoint \\ and timestamp with video interpolation and event emission model\end{tabular} &
  \cite{9812276} \\  
  \bottomrule
\end{tabular}%
}
\end{table}

\section{Simulating Reality: The Event-based Simulators}
\label{eventSimulators}

Event-based simulators are crucial for advancing event-based vision systems, providing synthetic data for algorithm validation and application exploration in a controlled, cost-effective manner.
Notable simulators include the DAVIS Simulator \cite{mueggler2017event}, which generated event streams, intensity frames, and depth maps with high temporal precision through time interpolation. The ESIM \cite{rebecq2018esim} extended this by offering an open-source platform for modeling camera motion in $3D$ scenes, producing events and comprehensive ground truth data.

The v2e simulator \cite{hu2021v2e} converted conventional video frames into realistic event-based data, addressing non-idealities such as Gaussian event threshold mismatch. The ICNS simulator \cite{joubert2021event } enhanced noise accuracy by integrating real pixel noise distributions.
The DVS-Voltmeter \cite{lin2022dvs} used a stochastic approach to simulate realistic events, incorporating voltage variations and noise effects from high-frame-rate videos. The V2CE Toolbox \cite{zhang2023v2ce} improved video-to-event conversion with dynamic-aware timestamp inference.
Additionally, the CARLA DVS camera \cite{carla_dvs2023} implementation simulates event generation with high-frequency execution to emulate microsecond resolution and adjust sensor frequency based on scene dynamics, while Prophesee Video to Event Simulator \cite{prophesee_event_simulator2023} provides a \texttt{Python} script for converting frame-based videos into event-based counterparts. Together, these simulators are essential for developing and testing event-based vision systems, driving innovation in the field. Tab. \ref{eventSimulators_table} summarizes the most commonly used event-based simulators. 


\section{Conclusion}
Event cameras have significantly impacted visual sensing technology and this survey outlines their evolution, explains their operational principles, and highlights how they differ from traditional frame-based cameras. It reviews various models and key milestones, offering a comprehensive overview of event-based vision as it stands today. The diverse applications of event cameras across different fields demonstrate their flexibility and potential. The importance of real-world and synthetic datasets in advancing the field is emphasized, along with the role of simulators in improving testing and development. As research progresses, consolidating and sharing knowledge will be essential for addressing new challenges and promoting further innovation. The \href{https://github.com/chakravarthi589/Event-based-Vision_Resources}{GitHub} page provided will be a valuable resource for the research community, offering access to past research and continuously updated with ongoing research and other relevant materials.


\clearpage  

%
%
\bibliographystyle{splncs04}
\bibliography{main}

\begin{thebibliography}{100}
\providecommand{\url}[1]{\texttt{#1}}
\providecommand{\urlprefix}{URL }
\providecommand{\doi}[1]{https://doi.org/#1}

\bibitem{ahmad2023person}
Ahmad, S., Morerio, P., Del~Bue, A.: Person re-identification without identification via event anonymization. In: Proceedings of the IEEE/CVF International Conference on Computer Vision. pp. 11132--11141 (2023)

\bibitem{aliminati2024sevd}
Aliminati, M.R., Chakravarthi, B., Verma, A.A., Vaghela, A., Wei, H., Zhou, X., Yang, Y.: Sevd: Synthetic event-based vision dataset for ego and fixed traffic perception. arXiv preprint arXiv:2404.10540  (2024)

\bibitem{9893571}
Alkendi, Y., Azzam, R., Ayyad, A., Javed, S., Seneviratne, L., Zweiri, Y.: Neuromorphic camera denoising using graph neural network-driven transformers. IEEE Transactions on Neural Networks and Learning Systems  \textbf{35}(3),  4110--4124 (2024). \doi{10.1109/TNNLS.2022.3201830}

\bibitem{alonso2019ev}
Alonso, I., Murillo, A.C.: Ev-segnet: Semantic segmentation for event-based cameras. In: Proceedings of the IEEE/CVF Conference on Computer Vision and Pattern Recognition Workshops. pp.~0--0 (2019)

\bibitem{9812276}
Amini, A., Wang, T.H., Gilitschenski, I., Schwarting, W., Liu, Z., Han, S., Karaman, S., Rus, D.: Vista 2.0: An open, data-driven simulator for multimodal sensing and policy learning for autonomous vehicles. In: 2022 International Conference on Robotics and Automation (ICRA). pp. 2419--2426 (2022). \doi{10.1109/ICRA46639.2022.9812276}

\bibitem{amir2017low}
Amir, A., Taba, B., Berg, D., Melano, T., McKinstry, J., Di~Nolfo, C., Nayak, T., Andreopoulos, A., Garreau, G., Mendoza, M., et~al.: A low power, fully event-based gesture recognition system. In: Proceedings of the IEEE conference on computer vision and pattern recognition. pp. 7243--7252 (2017)

\bibitem{andreopoulos2018low}
Andreopoulos, A., Kashyap, H.J., Nayak, T.K., Amir, A., Flickner, M.D.: A low power, high throughput, fully event-based stereo system. In: Proceedings of the IEEE conference on computer vision and pattern recognition. pp. 7532--7542 (2018)

\bibitem{bagchi2020event}
Bagchi, S., Chin, T.J.: Event-based star tracking via multiresolution progressive hough transforms. In: Proceedings of the IEEE/CVF Winter Conference on Applications of Computer Vision. pp. 2143--2152 (2020)

\bibitem{baldwin2020event}
Baldwin, R., Almatrafi, M., Asari, V., Hirakawa, K.: Event probability mask (epm) and event denoising convolutional neural network (edncnn) for neuromorphic cameras. In: Proceedings of the IEEE/CVF Conference on Computer Vision and Pattern Recognition. pp. 1701--1710 (2020)

\bibitem{bartolozzi2011embedded}
Bartolozzi, C., Rea, F., Clercq, C., Fasnacht, D.B., Indiveri, G., Hofst{\"a}tter, M., Metta, G.: Embedded neuromorphic vision for humanoid robots. In: CVPR 2011 workshops. pp. 129--135. IEEE (2011)

\bibitem{benosman2013event}
Benosman, R., Clercq, C., Lagorce, X., Ieng, S.H., Bartolozzi, C.: Event-based visual flow. IEEE transactions on neural networks and learning systems  \textbf{25}(2),  407--417 (2013)

\bibitem{bi2019graph}
Bi, Y., Chadha, A., Abbas, A., Bourtsoulatze, E., Andreopoulos, Y.: Graph-based object classification for neuromorphic vision sensing. In: Proceedings of the IEEE/CVF international conference on computer vision. pp. 491--501 (2019)

\bibitem{bi2020graph}
Bi, Y., Chadha, A., Abbas, A., Bourtsoulatze, E., Andreopoulos, Y.: Graph-based spatio-temporal feature learning for neuromorphic vision sensing. IEEE Transactions on Image Processing  \textbf{29},  9084--9098 (2020)

\bibitem{bichler2012extraction}
Bichler, O., Querlioz, D., Thorpe, S.J., Bourgoin, J.P., Gamrat, C.: Extraction of temporally correlated features from dynamic vision sensors with spike-timing-dependent plasticity. Neural networks  \textbf{32},  339--348 (2012)

\bibitem{binas2017ddd17}
Binas, J., Neil, D., Liu, S.C., Delbruck, T.: Ddd17: End-to-end davis driving dataset. arXiv preprint arXiv:1711.01458  (2017)

\bibitem{biswas2024halsie}
Biswas, S.D., Kosta, A., Liyanagedera, C., Apolinario, M., Roy, K.: Halsie: Hybrid approach to learning segmentation by simultaneously exploiting image and event modalities. In: 2024 IEEE/CVF Winter Conference on Applications of Computer Vision (WACV). pp. 5952--5962. IEEE (2024)

\bibitem{boretti2023pedro}
Boretti, C., Bich, P., Pareschi, F., Prono, L., Rovatti, R., Setti, G.: Pedro: An event-based dataset for person detection in robotics. In: Proceedings of the IEEE/CVF Conference on Computer Vision and Pattern Recognition. pp. 4065--4070 (2023)

\bibitem{brebion2023learning}
Brebion, V., Moreau, J., Davoine, F.: Learning to estimate two dense depths from lidar and event data. In: Scandinavian Conference on Image Analysis. pp. 517--533. Springer (2023)

\bibitem{bulthoff2003biologically}
B{\"u}lthoff, H.H., Lee, S.W., Poggio, T., Wallraven, C.: Biologically Motivated Computer Vision: Second International Workshop, BMCV 2002, T{\"u}bingen, Germany, November 22-24, 2002, Proceedings, vol.~2525. Springer (2003)

\bibitem{burner2022evimo2}
Burner, L., Mitrokhin, A., Ferm{\"u}ller, C., Aloimonos, Y.: Evimo2: an event camera dataset for motion segmentation, optical flow, structure from motion, and visual inertial odometry in indoor scenes with monocular or stereo algorithms. arXiv preprint arXiv:2205.03467  (2022)

\bibitem{camunas2011event}
Camunas-Mesa, L., Zamarreno-Ramos, C., Linares-Barranco, A., Acosta-Jimenez, A.J., Serrano-Gotarredona, T., Linares-Barranco, B.: An event-driven multi-kernel convolution processor module for event-driven vision sensors. IEEE Journal of Solid-State Circuits  \textbf{47}(2),  504--517 (2011)

\bibitem{cannici2019attention}
Cannici, M., Ciccone, M., Romanoni, A., Matteucci, M.: Attention mechanisms for object recognition with event-based cameras. In: 2019 IEEE Winter Conference on Applications of Computer Vision (WACV). pp. 1127--1136. IEEE (2019)

\bibitem{cannici2024mitigating}
Cannici, M., Scaramuzza, D.: Mitigating motion blur in neural radiance fields with events and frames. In: Proceedings of the IEEE/CVF Conference on Computer Vision and Pattern Recognition. pp. 9286--9296 (2024)

\bibitem{cao2023event}
Cao, C., Fu, X., Liu, H., Huang, Y., Wang, K., Luo, J., Zha, Z.J.: Event-guided person re-identification via sparse-dense complementary learning. In: Proceedings of the IEEE/CVF Conference on Computer Vision and Pattern Recognition. pp. 17990--17999 (2023)

\bibitem{censi2014low}
Censi, A., Scaramuzza, D.: Low-latency event-based visual odometry. In: 2014 IEEE International Conference on Robotics and Automation (ICRA). pp. 703--710. IEEE (2014)

\bibitem{chakravarthi2023event}
Chakravarthi, B., Manoj~Kumar, M., Pavan~Kumar, B.: Event-based sensing for improved traffic detection and tracking in intelligent transport systems toward sustainable mobility. In: International Conference on Interdisciplinary Approaches in Civil Engineering for Sustainable Development. pp. 83--95. Springer (2023)

\bibitem{9810191}
Chamorro, W., Solà, J., Andrade-Cetto, J.: Event-based line slam in real-time. IEEE Robotics and Automation Letters  \textbf{7}(3),  8146--8153 (2022). \doi{10.1109/LRA.2022.3187266}

\bibitem{chen2020event}
Chen, G., Cao, H., Conradt, J., Tang, H., Rohrbein, F., Knoll, A.: Event-based neuromorphic vision for autonomous driving: A paradigm shift for bio-inspired visual sensing and perception. IEEE Signal Processing Magazine  \textbf{37}(4),  34--49 (2020)

\bibitem{chen2024timerewind}
Chen, J., Feng, B.Y., Cai, H., Xie, M., Metzler, C., Ferm\"uller, C., Aloimonos, Y.: Timerewind: Rewinding time with image-and-events video diffusion. arXiv preprint arXiv:2403.13800  (2024)

\bibitem{chen2024segment}
Chen, Z., Zhu, Z., Zhang, Y., Hou, J., Shi, G., Wu, J.: Segment any event streams via weighted adaptation of pivotal tokens. In: Proceedings of the IEEE/CVF Conference on Computer Vision and Pattern Recognition. pp. 3890--3900 (2024)

\bibitem{keypoint_21_cvprw}
Chiberre, P., Perot, E., Sironi, A., Lepetit, V.: Detecting stable keypoints from events through image gradient prediction. In: 2021 IEEE/CVF Conference on Computer Vision and Pattern Recognition Workshops (CVPRW). pp. 1387--1394 (2021). \doi{10.1109/CVPRW53098.2021.00153}

\bibitem{cho2023learning}
Cho, H., Cho, J., Yoon, K.J.: Learning adaptive dense event stereo from the image domain. In: Proceedings of the IEEE/CVF Conference on Computer Vision and Pattern Recognition. pp. 17797--17807 (2023)

\bibitem{cho2023non}
Cho, H., Jeong, Y., Kim, T., Yoon, K.J.: Non-coaxial event-guided motion deblurring with spatial alignment. In: Proceedings of the IEEE/CVF International Conference on Computer Vision. pp. 12492--12503 (2023)

\bibitem{cho2023label}
Cho, H., Kim, H., Chae, Y., Yoon, K.J.: Label-free event-based object recognition via joint learning with image reconstruction from events. In: Proceedings of the IEEE/CVF International Conference on Computer Vision. pp. 19866--19877 (2023)

\bibitem{cho2024tta}
Cho, H., Kim, T., Jeong, Y., Yoon, K.J.: Tta-evf: Test-time adaptation for event-based video frame interpolation via reliable pixel and sample estimation. In: Proceedings of the IEEE/CVF Conference on Computer Vision and Pattern Recognition. pp. 25701--25711 (2024)

\bibitem{cho2022selection}
Cho, H., Yoon, K.J.: Selection and cross similarity for event-image deep stereo. In: European Conference on Computer Vision. pp. 470--486. Springer (2022)

\bibitem{cocheteux2024muli}
Cocheteux, M., Moreau, J., Davoine, F.: Muli-ev: Maintaining unperturbed lidar-event calibration. In: Proceedings of the IEEE/CVF Conference on Computer Vision and Pattern Recognition. pp. 4579--4586 (2024)

\bibitem{conradt2009embedded}
Conradt, J., Berner, R., Cook, M., Delbruck, T.: An embedded aer dynamic vision sensor for low-latency pole balancing. In: 2009 IEEE 12th International Conference on Computer Vision Workshops, ICCV Workshops. pp. 780--785. IEEE (2009)

\bibitem{conradt2009pencil}
Conradt, J., Cook, M., Berner, R., Lichtsteiner, P., Douglas, R.J., Delbruck, T.: A pencil balancing robot using a pair of aer dynamic vision sensors. In: 2009 IEEE International Symposium on Circuits and Systems. pp. 781--784. IEEE (2009)

\bibitem{Dai_2023_CVPR}
Dai, P., Zhang, Y., Yu, X., Lyu, X., Qi, X.: Hybrid neural rendering for large-scale scenes with motion blur. In: Proceedings of the IEEE/CVF Conference on Computer Vision and Pattern Recognition (CVPR). pp. 154--164 (June 2023)

\bibitem{de2020large}
De~Tournemire, P., Nitti, D., Perot, E., Migliore, D., Sironi, A.: A large scale event-based detection dataset for automotive. arXiv preprint arXiv:2001.08499  (2020)

\bibitem{delbruck2013robotic}
Delbruck, T., Lang, M.: Robotic goalie with 3 ms reaction time at 4\% cpu load using event-based dynamic vision sensor. Frontiers in neuroscience  \textbf{7}, ~223 (2013)

\bibitem{delbruck2010activity}
Delbr{\"u}ck, T., Linares-Barranco, B., Culurciello, E., Posch, C.: Activity-driven, event-based vision sensors. In: Proceedings of 2010 IEEE international symposium on circuits and systems. pp. 2426--2429. IEEE (2010)

\bibitem{delbruck2008frame}
Delbruck, T., et~al.: Frame-free dynamic digital vision. In: Proceedings of Intl. Symp. on Secure-Life Electronics, Advanced Electronics for Quality Life and Society. vol.~1, pp. 21--26. Citeseer (2008)

\bibitem{deng2022voxel}
Deng, Y., Chen, H., Liu, H., Li, Y.: A voxel graph cnn for object classification with event cameras. In: Proceedings of the IEEE/CVF Conference on Computer Vision and Pattern Recognition. pp. 1172--1181 (2022)

\bibitem{deniz2023neuromorphic}
Deniz, D., Ros, E., Ferm\"uller, C., Barranco, F.: When do neuromorphic sensors outperform cameras? learning from dynamic features. In: 2023 57th Annual Conference on Information Sciences and Systems (CISS). pp.~1--6. IEEE (2023)

\bibitem{dowling2005artificial}
Dowling, J.: Artificial human vision. Expert Review of Medical Devices  \textbf{2}(1),  73--85 (2005)

\bibitem{duan2021eventzoom}
Duan, P., Wang, Z.W., Zhou, X., Ma, Y., Shi, B.: Eventzoom: Learning to denoise and super resolve neuromorphic events. In: Proceedings of the IEEE/CVF conference on computer vision and pattern recognition. pp. 12824--12833 (2021)

\bibitem{duan2024led}
Duan, Y.: Led: A large-scale real-world paired dataset for event camera denoising. In: Proceedings of the IEEE/CVF Conference on Computer Vision and Pattern Recognition. pp. 25637--25647 (2024)

\bibitem{v2ce2023}
DVS, U.H.: V2ce toolbox (2023), \url{https://github.com/ucsd-hdsi-dvs/V2CE-Toolbox}, accessed: 2024-07-17

\bibitem{elms2024event}
Elms, E., Latif, Y., Park, T.H., Chin, T.J.: Event-based structure-from-orbit. In: Proceedings of the IEEE/CVF Conference on Computer Vision and Pattern Recognition. pp. 19541--19550 (2024)

\bibitem{EvShutter}
Erbach, J., Tulyakov, S., Vitoria, P., Bochicchio, A., Li, Y.: Evshutter: Transforming events for unconstrained rolling shutter correction. In: 2023 IEEE/CVF Conference on Computer Vision and Pattern Recognition (CVPR). pp. 13904--13913 (2023). \doi{10.1109/CVPR52729.2023.01336}

\bibitem{10462903}
Ercan, B., Eker, O., Saglam, C., Erdem, A., Erdem, E.: Hypere2vid: Improving event-based video reconstruction via hypernetworks. IEEE Transactions on Image Processing  \textbf{33},  1826--1837 (2024). \doi{10.1109/TIP.2024.3372460}

\bibitem{everding2018low}
Everding, L., Conradt, J.: Low-latency line tracking using event-based dynamic vision sensors. Frontiers in neurorobotics  \textbf{12}, ~4 (2018)

\bibitem{falanga2020dynamic}
Falanga, D., Kleber, K., Scaramuzza, D.: Dynamic obstacle avoidance for quadrotors with event cameras. Science Robotics  \textbf{5}(40),  eaaz9712 (2020)

\bibitem{firouzi2016asynchronous}
Firouzi, M., Conradt, J.: Asynchronous event-based cooperative stereo matching using neuromorphic silicon retinas. Neural Processing Letters  \textbf{43},  311--326 (2016)

\bibitem{floreano2001evolution}
Floreano, D., Mattiussi, C.: Evolution of spiking neural controllers for autonomous vision-based robots. In: International Symposium on Evolutionary Robotics. pp. 38--61. Springer (2001)

\bibitem{10161392}
Forrai, B., Miki, T., Gehrig, D., Hutter, M., Scaramuzza, D.: Event-based agile object catching with a quadrupedal robot. In: 2023 IEEE International Conference on Robotics and Automation (ICRA). pp. 12177--12183 (2023). \doi{10.1109/ICRA48891.2023.10161392}

\bibitem{davis_simulator2023}
Forster, C., Liu, H., Chen, Q., Scaramuzza, D.: Davis simulator. \url{https://github.com/uzh-rpg/rpg_davis_simulator} (2023)

\bibitem{esim2023}
Forster, C., Liu, H., Chen, Q., Scaramuzza, D.: Esim: an open event camera simulator. \url{https://github.com/uzh-rpg/rpg_esim} (2023)

\bibitem{gallego2020event}
Gallego, G., Delbr{\"u}ck, T., Orchard, G., Bartolozzi, C., Taba, B., Censi, A., Leutenegger, S., Davison, A.J., Conradt, J., Daniilidis, K., et~al.: Event-based vision: A survey. IEEE transactions on pattern analysis and machine intelligence  \textbf{44}(1),  154--180 (2020)

\bibitem{gallego2019focus}
Gallego, G., Gehrig, M., Scaramuzza, D.: Focus is all you need: Loss functions for event-based vision. In: Proceedings of the IEEE/CVF Conference on Computer Vision and Pattern Recognition. pp. 12280--12289 (2019)

\bibitem{gallego2018unifying}
Gallego, G., Rebecq, H., Scaramuzza, D.: A unifying contrast maximization framework for event cameras, with applications to motion, depth, and optical flow estimation. In: Proceedings of the IEEE conference on computer vision and pattern recognition. pp. 3867--3876 (2018)

\bibitem{gao2024n}
Gao, L., Gehrig, D., Su, H., Scaramuzza, D., Kneip, L.: An n-point linear solver for line and motion estimation with event cameras. In: Proceedings of the IEEE/CVF Conference on Computer Vision and Pattern Recognition. pp. 14596--14605 (2024)

\bibitem{9809788}
Gao, L., Liang, Y., Yang, J., Wu, S., Wang, C., Chen, J., Kneip, L.: Vector: A versatile event-centric benchmark for multi-sensor slam. IEEE Robotics and Automation Letters  \textbf{7}(3),  8217--8224 (2022). \doi{10.1109/LRA.2022.3186770}

\bibitem{gao2024sd2event}
Gao, Y., Zhu, Y., Li, X., Du, Y., Zhang, T.: Sd2event: Self-supervised learning of dynamic detectors and contextual descriptors for event cameras. In: Proceedings of the IEEE/CVF Conference on Computer Vision and Pattern Recognition. pp. 3055--3064 (2024)

\bibitem{10480584}
Gao, Y., Lu, J., Li, S., Li, Y., Du, S.: Hypergraph-based multi-view action recognition using event cameras. IEEE Transactions on Pattern Analysis and Machine Intelligence pp. 1--14 (2024). \doi{10.1109/TPAMI.2024.3382117}

\bibitem{gehrig2020video}
Gehrig, D., Gehrig, M., Hidalgo-Carri{\'o}, J., Scaramuzza, D.: Video to events: Recycling video datasets for event cameras. In: Proceedings of the IEEE/CVF Conference on Computer Vision and Pattern Recognition. pp. 3586--3595 (2020)

\bibitem{gehrig2018asynchronous}
Gehrig, D., Rebecq, H., Gallego, G., Scaramuzza, D.: Asynchronous, photometric feature tracking using events and frames. In: Proceedings of the European Conference on Computer Vision (ECCV). pp. 750--765 (2018)

\bibitem{Gehrig24nature_DAGr}
Gehrig, D., Scaramuzza, D.: Low latency automotive vision with event cameras  (2024)

\bibitem{gehrig2021dsec}
Gehrig, M., Aarents, W., Gehrig, D., Scaramuzza, D.: Dsec: A stereo event camera dataset for driving scenarios. IEEE Robotics and Automation Letters  \textbf{6}(3),  4947--4954 (2021)

\bibitem{gehrig2023recurrent}
Gehrig, M., Scaramuzza, D.: Recurrent vision transformers for object detection with event cameras. In: Proceedings of the IEEE/CVF conference on computer vision and pattern recognition. pp. 13884--13893 (2023)

\bibitem{geng2024event}
Geng, M., Zhu, L., Wang, L., Zhang, W., Xiong, R., Tian, Y.: Event-based visible and infrared fusion via multi-task collaboration. In: Proceedings of the IEEE/CVF Conference on Computer Vision and Pattern Recognition. pp. 26929--26939 (2024)

\bibitem{10160768}
Gentil, C.L., Alzugaray, I., Vidal-Calleja, T.: Continuous-time gaussian process motion-compensation for event-vision pattern tracking with distance fields. In: 2023 IEEE International Conference on Robotics and Automation (ICRA). pp. 804--812 (2023). \doi{10.1109/ICRA48891.2023.10160768}

\bibitem{gokarn2024poster}
Gokarn, I., Misra, A.: Poster: Profiling event vision processing on edge devices. In: Proceedings of the 22nd Annual International Conference on Mobile Systems, Applications and Services. pp. 672--673 (2024)

\bibitem{9720086}
Guo, S., Delbruck, T.: Low cost and latency event camera background activity denoising. IEEE Transactions on Pattern Analysis and Machine Intelligence  \textbf{45}(1),  785--795 (2023). \doi{10.1109/TPAMI.2022.3152999}

\bibitem{Guo24tro}
Guo, S., Gallego, G.: {CMax}-{SLAM}: Event-based rotational-motion bundle adjustment and {SLAM} system using contrast maximization. {IEEE} Transactions on Robotics  \textbf{40},  2442--2461 (2024). \doi{10.1109/TRO.2024.3378443}

\bibitem{habuchi2024time}
Habuchi, S., Takahashi, K., Tsutake, C., Fujii, T., Nagahara, H.: Time-efficient light-field acquisition using coded aperture and events. In: Proceedings of the IEEE/CVF Conference on Computer Vision and Pattern Recognition. pp. 24923--24933 (2024)

\bibitem{Han_2023_CVPR}
Han, J., Asano, Y., Shi, B., Zheng, Y., Sato, I.: High-fidelity event-radiance recovery via transient event frequency. In: Proceedings of the IEEE/CVF Conference on Computer Vision and Pattern Recognition (CVPR). pp. 20616--20625 (June 2023)

\bibitem{han2021evintsr}
Han, J., Yang, Y., Zhou, C., Xu, C., Shi, B.: Evintsr-net: Event guided multiple latent frames reconstruction and super-resolution. In: Proceedings of the IEEE/CVF International Conference on Computer Vision. pp. 4882--4891 (2021)

\bibitem{harrison1998neuromorphic}
Harrison, R.R., Koch, C.: A neuromorphic visual motion sensor for real-world robots. In: Workshop on Defining the Future of Biomorphic Robotics, IROS. vol.~98. Citeseer (1998)

\bibitem{he2022timereplayer}
He, W., You, K., Qiao, Z., Jia, X., Zhang, Z., Wang, W., Lu, H., Wang, Y., Liao, J.: Timereplayer: Unlocking the potential of event cameras for video interpolation. In: Proceedings of the IEEE/CVF Conference on Computer Vision and Pattern Recognition. pp. 17804--17813 (2022)

\bibitem{hidalgo2022event}
Hidalgo-Carri{\'o}, J., Gallego, G., Scaramuzza, D.: Event-aided direct sparse odometry. In: Proceedings of the IEEE/CVF Conference on Computer Vision and Pattern Recognition. pp. 5781--5790 (2022)

\bibitem{eCDT_22_IROS}
Hu, S., Kim, Y., Lim, H., Lee, A.J., Myung, H.: ecdt: Event clustering for simultaneous feature detection and tracking. In: 2022 IEEE/RSJ International Conference on Intelligent Robots and Systems (IROS). pp. 3808--3815 (2022). \doi{10.1109/IROS47612.2022.9981451}

\bibitem{hu2021v2e}
Hu, Y., Liu, S.C., Delbruck, T.: v2e: From video frames to realistic dvs events. In: Proceedings of the IEEE/CVF conference on computer vision and pattern recognition. pp. 1312--1321 (2021)

\bibitem{huang2023progressive}
Huang, X., Zhang, Y., Xiong, Z.: Progressive spatio-temporal alignment for efficient event-based motion estimation. In: Proceedings of the IEEE/CVF Conference on Computer Vision and Pattern Recognition. pp. 1537--1546 (2023)

\bibitem{huang2023eventpoint}
Huang, Z., Sun, L., Zhao, C., Li, S., Su, S.: Eventpoint: Self-supervised interest point detection and description for event-based camera. In: Proceedings of the IEEE/CVF Winter Conference on Applications of Computer Vision. pp. 5396--5405 (2023)

\bibitem{huang2024bilateral}
Huang, Z., Liang, Q., Yu, Y., Qin, C., Zheng, X., Huang, K., Zhou, Z., Yang, W.: Bilateral event mining and complementary for event stream super-resolution. In: Proceedings of the IEEE/CVF Conference on Computer Vision and Pattern Recognition. pp. 34--43 (2024)

\bibitem{hwang2023ev}
Hwang, I., Kim, J., Kim, Y.M.: Ev-nerf: Event based neural radiance field. In: Proceedings of the IEEE/CVF Winter Conference on Applications of Computer Vision. pp. 837--847 (2023)

\bibitem{iaboni2022event}
Iaboni, C., Lobo, D., Choi, J.W., Abichandani, P.: Event-based motion capture system for online multi-quadrotor localization and tracking. Sensors  \textbf{22}(9), ~3240 (2022)

\bibitem{imatestDynamicRange}
{Imatest}: Dynamic range. \url{https://www.imatest.com/solutions/dynamic-range/} (2024), accessed: 2024-07-23

\bibitem{inivation_davis240}
iniVation: Davis 240 dvs event camera (2019), \url{https://inivation.com/wp-content/uploads/2019/08/DAVIS240.pdf}, accessed: 2024-07-18

\bibitem{inivation_dvs128}
iniVation: Dvs128 event camera (2019), \url{https://inivation.com/wp-content/uploads/2019/08/DVS128.pdf}, accessed: 2024-07-18

\bibitem{davis_346_aer_event_camera}
iniVation: Davis 346 aer - event camera (2024), \url{https://docs.inivation.com/hardware/current-products/davis346-aer.html}, accessed: 2024-07-18

\bibitem{dv_software}
iniVation: Dv - open-source software (2024), \url{https://docs.inivation.com/software/dv/index.html}, accessed: 2024-07-18

\bibitem{dv_processing}
iniVation: Dv-processing - c++/python library (2024), \url{https://docs.inivation.com/software/dv-processing.html}, accessed: 2024-07-18

\bibitem{davis346_event_camera}
iniVation: Dvs 346 event camera (2024), \url{https://docs.inivation.com/hardware/current-products/davis346.html}, accessed: 2024-07-18

\bibitem{dvxplorer_event_camera}
iniVation: Dvxplorer - event camera (2024), \url{https://docs.inivation.com/hardware/current-products/dvxplorer.html}, accessed: 2024-07-18

\bibitem{dvxplorer_lite_event_camera}
iniVation: Dvxplorer lite - event camera (2024), \url{https://docs.inivation.com/hardware/current-products/dvxplorer-lite.html}, accessed: 2024-07-18

\bibitem{dvxplorer_s_duo_event_camera}
iniVation: Dvxplorer s duo (2024), \url{https://docs.inivation.com/hardware/current-products/dvxplorer-s-duo.html}, accessed: 2024-07-18

\bibitem{inivation_website}
iniVation: inivation – neuromorphic vision systems (2024), \url{https://inivation.com/}, accessed: 2024-07-18

\bibitem{stereo_kit}
iniVation: Stereo kit (2024), \url{https://docs.inivation.com/hardware/current-products/stereo-kit.html}, accessed: 2024-07-18

\bibitem{insightness_website}
Insightness: Insightness – sight for your device (2024), \url{https://www.insightness.com/}, accessed: 2024-07-18

\bibitem{itti2004ilab}
Itti, L.: The ilab neuromorphic vision c++ toolkit: Free tools for the next generation of vision algorithms. The Neuromorphic Engineer  \textbf{1}(1), ~10 (2004)

\bibitem{10160531}
Jawaid, M., Elms, E., Latif, Y., Chin, T.J.: Towards bridging the space domain gap for satellite pose estimation using event sensing. In: 2023 IEEE International Conference on Robotics and Automation (ICRA). pp. 11866--11873 (2023). \doi{10.1109/ICRA48891.2023.10160531}

\bibitem{jiang2024complementing}
Jiang, J., Zhou, X., Wang, B., Deng, X., Xu, C., Shi, B.: Complementing event streams and rgb frames for hand mesh reconstruction. In: Proceedings of the IEEE/CVF Conference on Computer Vision and Pattern Recognition. pp. 24944--24954 (2024)

\bibitem{jiang2020learning}
Jiang, Z., Zhang, Y., Zou, D., Ren, J., Lv, J., Liu, Y.: Learning event-based motion deblurring. In: Proceedings of the IEEE/CVF Conference on Computer Vision and Pattern Recognition. pp. 3320--3329 (2020)

\bibitem{jiao2021comparing}
Jiao, J., Huang, H., Li, L., He, Z., Zhu, Y., Liu, M.: Comparing representations in tracking for event camera-based slam. In: Proceedings of the IEEE/cvf conference on computer vision and pattern recognition. pp. 1369--1376 (2021)

\bibitem{jing2024hpl}
Jing, L., Ding, Y., Gao, Y., Wang, Z., Yan, X., Wang, D., Schaefer, G., Fang, H., Zhao, B., Li, X.: Hpl-ess: Hybrid pseudo-labeling for unsupervised event-based semantic segmentation. In: Proceedings of the IEEE/CVF Conference on Computer Vision and Pattern Recognition. pp. 23128--23137 (2024)

\bibitem{jing2021turning}
Jing, Y., Yang, Y., Wang, X., Song, M., Tao, D.: Turning frequency to resolution: Video super-resolution via event cameras. In: Proceedings of the IEEE/CVF Conference on Computer Vision and Pattern Recognition. pp. 7772--7781 (2021)

\bibitem{joubert2021event}
Joubert, D., Marcireau, A., Ralph, N., Jolley, A., Van~Schaik, A., Cohen, G.: Event camera simulator improvements via characterized parameters. Frontiers in Neuroscience  \textbf{15},  702765 (2021)

\bibitem{khodamoradi2018n}
Khodamoradi, A., Kastner, R.: $ o (n) $ o (n)-space spatiotemporal filter for reducing noise in neuromorphic vision sensors. IEEE Transactions on Emerging Topics in Computing  \textbf{9}(1),  15--23 (2018)

\bibitem{kim2021n}
Kim, J., Bae, J., Park, G., Zhang, D., Kim, Y.M.: N-imagenet: Towards robust, fine-grained object recognition with event cameras. In: Proceedings of the IEEE/CVF international conference on computer vision. pp. 2146--2156 (2021)

\bibitem{kim2022ev}
Kim, J., Hwang, I., Kim, Y.M.: Ev-tta: Test-time adaptation for event-based object recognition. In: Proceedings of the IEEE/CVF Conference on Computer Vision and Pattern Recognition. pp. 17745--17754 (2022)

\bibitem{kim2023event}
Kim, T., Chae, Y., Jang, H.K., Yoon, K.J.: Event-based video frame interpolation with cross-modal asymmetric bidirectional motion fields. In: Proceedings of the IEEE/CVF Conference on Computer Vision and Pattern Recognition. pp. 18032--18042 (2023)

\bibitem{kim2024frequency}
Kim, T., Cho, H., Yoon, K.J.: Frequency-aware event-based video deblurring for real-world motion blur. In: Proceedings of the IEEE/CVF Conference on Computer Vision and Pattern Recognition. pp. 24966--24976 (2024)

\bibitem{kim2022event}
Kim, T., Lee, J., Wang, L., Yoon, K.J.: Event-guided deblurring of unknown exposure time videos. In: European Conference on Computer Vision. pp. 519--538. Springer (2022)

\bibitem{klenk2024masked}
Klenk, S., Bonello, D., Koestler, L., Araslanov, N., Cremers, D.: Masked event modeling: Self-supervised pretraining for event cameras. In: Proceedings of the IEEE/CVF Winter Conference on Applications of Computer Vision. pp. 2378--2388 (2024)

\bibitem{klenk2023nerf}
Klenk, S., Koestler, L., Scaramuzza, D., Cremers, D.: E-nerf: Neural radiance fields from a moving event camera. IEEE Robotics and Automation Letters  \textbf{8}(3),  1587--1594 (2023)

\bibitem{10067520}
Kodama, K., Sato, Y., Yorikado, Y., Berner, R., Mizoguchi, K., Miyazaki, T., Tsukamoto, M., Matoba, Y., Shinozaki, H., Niwa, A., Yamaguchi, T., Brandli, C., Wakabayashi, H., Oike, Y.: 1.22μm 35.6mpixel rgb hybrid event-based vision sensor with 4.88μm-pitch event pixels and up to 10k event frame rate by adaptive control on event sparsity. In: 2023 IEEE International Solid-State Circuits Conference (ISSCC). pp. 92--94 (2023). \doi{10.1109/ISSCC42615.2023.10067520}

\bibitem{kong2024openess}
Kong, L., Liu, Y., Ng, L.X., Cottereau, B.R., Ooi, W.T.: Openess: Event-based semantic scene understanding with open vocabularies. In: Proceedings of the IEEE/CVF Conference on Computer Vision and Pattern Recognition. pp. 15686--15698 (2024)

\bibitem{10160551}
Kosta, A.K., Roy, K.: Adaptive-spikenet: Event-based optical flow estimation using spiking neural networks with learnable neuronal dynamics. In: 2023 IEEE International Conference on Robotics and Automation (ICRA). pp. 6021--6027 (2023). \doi{10.1109/ICRA48891.2023.10160551}

\bibitem{10160276}
Kılıç, O.S., Akman, A., Alatan, A.A.: E-vfia: Event-based video frame interpolation with attention. In: 2023 IEEE International Conference on Robotics and Automation (ICRA). pp. 8284--8290 (2023). \doi{10.1109/ICRA48891.2023.10160276}

\bibitem{lucid_website}
Labs, L.V.: Lucid vision labs: Industrial machine vision cameras (2024), \url{https://thinklucid.com/}, accessed: 2024-07-18

\bibitem{lagorce2014asynchronous}
Lagorce, X., Meyer, C., Ieng, S.H., Filliat, D., Benosman, R.: Asynchronous event-based multikernel algorithm for high-speed visual features tracking. IEEE transactions on neural networks and learning systems  \textbf{26}(8),  1710--1720 (2014)

\bibitem{lee2020spike}
Lee, C., Kosta, A.K., Zhu, A.Z., Chaney, K., Daniilidis, K., Roy, K.: Spike-flownet: event-based optical flow estimation with energy-efficient hybrid neural networks. In: European Conference on Computer Vision. pp. 366--382. Springer (2020)

\bibitem{10161098}
Lee, M.S., Kim, Y.J., Jung, J.H., Park, C.G.: Fusion of events and frames using 8-dof warping model for robust feature tracking. In: 2023 IEEE International Conference on Robotics and Automation (ICRA). pp. 834--840 (2023). \doi{10.1109/ICRA48891.2023.10161098}

\bibitem{li2024event}
Li, H., Wang, J., Yuan, J., Li, Y., Weng, W., Peng, Y., Zhang, Y., Xiong, Z., Sun, X.: Event-assisted low-light video object segmentation. In: Proceedings of the IEEE/CVF Conference on Computer Vision and Pattern Recognition. pp. 3250--3259 (2024)

\bibitem{li2017cifar10}
Li, H., Liu, H., Ji, X., Li, G., Shi, L.: Cifar10-dvs: an event-stream dataset for object classification. Frontiers in neuroscience  \textbf{11}, ~309 (2017)

\bibitem{ASTM-NET}
Li, J., Li, J., Zhu, L., Xiang, X., Huang, T., Tian, Y.: Asynchronous spatio-temporal memory network for continuous event-based object detection. IEEE Transactions on Image Processing  \textbf{31},  2975--2987 (2022). \doi{10.1109/TIP.2022.3162962}

\bibitem{li2021event}
Li, S., Feng, Y., Li, Y., Jiang, Y., Zou, C., Gao, Y.: Event stream super-resolution via spatiotemporal constraint learning. In: Proceedings of the IEEE/CVF International Conference on Computer Vision. pp. 4480--4489 (2021)

\bibitem{li20243d}
Li, S., Zhou, Z., Xue, Z., Li, Y., Du, S., Gao, Y.: 3d feature tracking via event camera. In: Proceedings of the IEEE/CVF Conference on Computer Vision and Pattern Recognition. pp. 18974--18983 (2024)

\bibitem{li2021graph}
Li, Y., Zhou, H., Yang, B., Zhang, Y., Cui, Z., Bao, H., Zhang, G.: Graph-based asynchronous event processing for rapid object recognition. In: Proceedings of the IEEE/CVF International Conference on Computer Vision. pp. 934--943 (2021)

\bibitem{liang2024towards}
Liang, G., Chen, K., Li, H., Lu, Y., Wang, L.: Towards robust event-guided low-light image enhancement: A large-scale real-world event-image dataset and novel approach. In: Proceedings of the IEEE/CVF Conference on Computer Vision and Pattern Recognition. pp. 23--33 (2024)

\bibitem{liang2023coherent}
Liang, J., Yang, Y., Li, B., Duan, P., Xu, Y., Shi, B.: Coherent event guided low-light video enhancement. In: Proceedings of the IEEE/CVF International Conference on Computer Vision. pp. 10615--10625 (2023)

\bibitem{9878722}
Liao, W., Zhang, X., Yu, L., Lin, S., Yang, W., Qiao, N.: Synthetic aperture imaging with events and frames. In: 2022 IEEE/CVF Conference on Computer Vision and Pattern Recognition (CVPR). pp. 17714--17723 (2022). \doi{10.1109/CVPR52688.2022.01721}

\bibitem{lichtsteiner2008128}
Lichtsteiner, P., Posch, C., Delbruck, T.: A 128x128 120 db 15 μs latency asynchronous temporal contrast vision sensor. IEEE journal of solid-state circuits  \textbf{43}(2),  566--576 (2008)

\bibitem{lin2022dvs}
Lin, S., Ma, Y., Guo, Z., Wen, B.: Dvs-voltmeter: Stochastic process-based event simulator for dynamic vision sensors. In: European Conference on Computer Vision. pp. 578--593. Springer (2022)

\bibitem{lin2020learning}
Lin, S., Zhang, J., Pan, J., Jiang, Z., Zou, D., Wang, Y., Chen, J., Ren, J.: Learning event-driven video deblurring and interpolation. In: Computer Vision--ECCV 2020: 16th European Conference, Glasgow, UK, August 23--28, 2020, Proceedings, Part VIII 16. pp. 695--710. Springer (2020)

\bibitem{litzenberger2006embedded}
Litzenberger, M., Posch, C., Bauer, D., Belbachir, A.N., Schon, P., Kohn, B., Garn, H.: Embedded vision system for real-time object tracking using an asynchronous transient vision sensor. In: 2006 IEEE 12th Digital Signal Processing Workshop \& 4th IEEE Signal Processing Education Workshop. pp. 173--178. IEEE (2006)

\bibitem{liu2022fast}
Liu, C., Qi, X., Lam, E.Y., Wong, N.: Fast classification and action recognition with event-based imaging. IEEE access  \textbf{10},  55638--55649 (2022)

\bibitem{liu2020globally}
Liu, D., Parra, A., Chin, T.J.: Globally optimal contrast maximisation for event-based motion estimation. In: Proceedings of the IEEE/CVF Conference on Computer Vision and Pattern Recognition. pp. 6349--6358 (2020)

\bibitem{liu2021spatiotemporal}
Liu, D., Parra, A., Chin, T.J.: Spatiotemporal registration for event-based visual odometry. In: Proceedings of the IEEE/CVF conference on computer vision and pattern recognition. pp. 4937--4946 (2021)

\bibitem{9811943}
Liu, D., Parra, A., Latif, Y., Chen, B., Chin, T.J., Reid, I.: Asynchronous optimisation for event-based visual odometry. In: 2022 International Conference on Robotics and Automation (ICRA). pp. 9432--9438 (2022). \doi{10.1109/ICRA46639.2022.9811943}

\bibitem{liu2017high}
Liu, H.C., Zhang, F.L., Marshall, D., Shi, L., Hu, S.M.: High-speed video generation with an event camera. The Visual Computer  \textbf{33},  749--759 (2017)

\bibitem{liu2023tma}
Liu, H., Chen, G., Qu, S., Zhang, Y., Li, Z., Knoll, A., Jiang, C.: Tma: Temporal motion aggregation for event-based optical flow. In: Proceedings of the IEEE/CVF International Conference on Computer Vision. pp. 9685--9694 (2023)

\bibitem{liu2024seeing}
Liu, H., Peng, S., Zhu, L., Chang, Y., Zhou, H., Yan, L.: Seeing motion at nighttime with an event camera. In: Proceedings of the IEEE/CVF Conference on Computer Vision and Pattern Recognition. pp. 25648--25658 (2024)

\bibitem{EDFLOW}
Liu, M., Delbruck, T.: Edflow: Event driven optical flow camera with keypoint detection and adaptive block matching. IEEE Transactions on Circuits and Systems for Video Technology  \textbf{32}, ~1--1 (09 2022). \doi{10.1109/TCSVT.2022.3156653}

\bibitem{9811687}
Liu, P., Chen, G., Li, Z., Tang, H., Knoll, A.: Learning local event-based descriptor for patch-based stereo matching. In: 2022 International Conference on Robotics and Automation (ICRA). pp. 412--418 (2022). \doi{10.1109/ICRA46639.2022.9811687}

\bibitem{liu2024video}
Liu, Y., Deng, Y., Chen, H., Yang, Z.: Video frame interpolation via direct synthesis with the event-based reference. In: Proceedings of the IEEE/CVF Conference on Computer Vision and Pattern Recognition. pp. 8477--8487 (2024)

\bibitem{Lou_2023_CVPR}
Lou, H., Teng, M., Yang, Y., Shi, B.: All-in-focus imaging from event focal stack. In: Proceedings of the IEEE/CVF Conference on Computer Vision and Pattern Recognition (CVPR). pp. 17366--17375 (June 2023)

\bibitem{low2023robust}
Low, W.F., Lee, G.H.: Robust e-nerf: Nerf from sparse \& noisy events under non-uniform motion. In: Proceedings of the IEEE/CVF International Conference on Computer Vision. pp. 18335--18346 (2023)

\bibitem{lu2023learning}
Lu, Y., Wang, Z., Liu, M., Wang, H., Wang, L.: Learning spatial-temporal implicit neural representations for event-guided video super-resolution. In: Proceedings of the IEEE/CVF Conference on Computer Vision and Pattern Recognition. pp. 1557--1567 (2023)

\bibitem{Luo_2024_CVPR}
Luo, X., Luo, A., Wang, Z., Lin, C., Zeng, B., Liu, S.: Efficient meshflow and optical flow estimation from event cameras. In: Proceedings of the IEEE/CVF Conference on Computer Vision and Pattern Recognition (CVPR). pp. 19198--19207 (June 2024)

\bibitem{luo2023learning}
Luo, X., Luo, K., Luo, A., Wang, Z., Tan, P., Liu, S.: Learning optical flow from event camera with rendered dataset. In: Proceedings of the IEEE/CVF International Conference on Computer Vision. pp. 9847--9857 (2023)

\bibitem{dvs_voltmeter2023}
Lynn: Dvs-voltmeter (2023), \url{https://github.com/Lynn0306/DVS-Voltmeter}, accessed: 2024-07-17

\bibitem{ma2023deformable}
Ma, Q., Paudel, D.P., Chhatkuli, A., Van~Gool, L.: Deformable neural radiance fields using rgb and event cameras. In: Proceedings of the IEEE/CVF International Conference on Computer Vision. pp. 3590--3600 (2023)

\bibitem{maashri2012accelerating}
Maashri, A.A., Debole, M., Cotter, M., Chandramoorthy, N., Xiao, Y., Narayanan, V., Chakrabarti, C.: Accelerating neuromorphic vision algorithms for recognition. In: Proceedings of the 49th annual design automation conference. pp. 579--584 (2012)

\bibitem{mahlknecht2022exploring}
Mahlknecht, F., Gehrig, D., Nash, J., Rockenbauer, F.M., Morrell, B., Delaune, J., Scaramuzza, D.: Exploring event camera-based odometry for planetary robots. IEEE Robotics and Automation Letters  \textbf{7}(4),  8651--8658 (2022)

\bibitem{manderscheid2019speed}
Manderscheid, J., Sironi, A., Bourdis, N., Migliore, D., Lepetit, V.: Speed invariant time surface for learning to detect corner points with event-based cameras. In: Proceedings of the IEEE/CVF Conference on Computer Vision and Pattern Recognition. pp. 10245--10254 (2019)

\bibitem{maqueda2018event}
Maqueda, A.I., Loquercio, A., Gallego, G., Garc{\'\i}a, N., Scaramuzza, D.: Event-based vision meets deep learning on steering prediction for self-driving cars. In: Proceedings of the IEEE conference on computer vision and pattern recognition. pp. 5419--5427 (2018)

\bibitem{9812142}
Masuda, M., Sekikawa, Y., Fujii, R., Saito, H.: Neural implicit event generator for motion tracking. In: 2022 International Conference on Robotics and Automation (ICRA). pp. 2200--2206 (2022). \doi{10.1109/ICRA46639.2022.9812142}

\bibitem{Mei_2023_CVPR}
Mei, H., Wang, Z., Yang, X., Wei, X., Delbruck, T.: Deep polarization reconstruction with pdavis events. In: Proceedings of the IEEE/CVF Conference on Computer Vision and Pattern Recognition (CVPR). pp. 22149--22158 (June 2023)

\bibitem{mesquida:cea-04321175_G2N2}
Mesquida, T., Dampfhoffer, M., Dalgaty, T., Vivet, P., Sironi, A., Posch, C.: {G2N2: Lightweight event stream classification with GRU graph neural networks}. In: {BMVC 2023 - The 34th British Machine Vision Conference}. p.~660. https://proceedings.bmvc2023.org/, Aberdeen, United Kingdom (Nov 2023), \url{https://cea.hal.science/cea-04321175}

\bibitem{messikommer2023data}
Messikommer, N., Fang, C., Gehrig, M., Scaramuzza, D.: Data-driven feature tracking for event cameras. In: Proceedings of the IEEE/CVF Conference on Computer Vision and Pattern Recognition. pp. 5642--5651 (2023)

\bibitem{millerdurai2024eventego3d}
Millerdurai, C., Akada, H., Wang, J., Luvizon, D., Theobalt, C., Golyanik, V.: Eventego3d: 3d human motion capture from egocentric event streams. In: Proceedings of the IEEE/CVF Conference on Computer Vision and Pattern Recognition. pp. 1186--1195 (2024)

\bibitem{mitrokhin2018event}
Mitrokhin, A., Ferm{\"u}ller, C., Parameshwara, C., Aloimonos, Y.: Event-based moving object detection and tracking. In: 2018 IEEE/RSJ International Conference on Intelligent Robots and Systems (IROS). pp.~1--9. IEEE (2018)

\bibitem{mitrokhin2020learning}
Mitrokhin, A., Hua, Z., Fermuller, C., Aloimonos, Y.: Learning visual motion segmentation using event surfaces. In: Proceedings of the IEEE/CVF Conference on Computer Vision and Pattern Recognition. pp. 14414--14423 (2020)

\bibitem{mitrokhin2019ev}
Mitrokhin, A., Ye, C., Ferm{\"u}ller, C., Aloimonos, Y., Delbruck, T.: Ev-imo: Motion segmentation dataset and learning pipeline for event cameras. In: 2019 IEEE/RSJ International Conference on Intelligent Robots and Systems (IROS). pp. 6105--6112. IEEE (2019)

\bibitem{10254473}
Mollica, G., Felicioni, S., Legittimo, M., Meli, L., Costante, G., Valigi, P.: Ma-vied: A multisensor automotive visual inertial event dataset. IEEE Transactions on Intelligent Transportation Systems  \textbf{25}(1),  214--224 (2024). \doi{10.1109/TITS.2023.3312355}

\bibitem{mostafavi2021event}
Mostafavi, M., Yoon, K.J., Choi, J.: Event-intensity stereo: Estimating depth by the best of both worlds. In: Proceedings of the IEEE/CVF International Conference on Computer Vision. pp. 4258--4267 (2021)

\bibitem{mueggler2015lifetime}
Mueggler, E., Forster, C., Baumli, N., Gallego, G., Scaramuzza, D.: Lifetime estimation of events from dynamic vision sensors. In: 2015 IEEE international conference on Robotics and Automation (ICRA). pp. 4874--4881. IEEE (2015)

\bibitem{mueggler2015continuous}
Mueggler, E., Gallego, G., Scaramuzza, D.: Continuous-time trajectory estimation for event-based vision sensors. In: Robotics: Science and Systems XI (2015)

\bibitem{mueggler2014event}
Mueggler, E., Huber, B., Scaramuzza, D.: Event-based, 6-dof pose tracking for high-speed maneuvers. In: 2014 IEEE/RSJ International Conference on Intelligent Robots and Systems. pp. 2761--2768. IEEE (2014)

\bibitem{mueggler2017event}
Mueggler, E., Rebecq, H., Gallego, G., Delbruck, T., Scaramuzza, D.: The event-camera dataset and simulator: Event-based data for pose estimation, visual odometry, and slam. The International Journal of Robotics Research  \textbf{36}(2),  142--149 (2017)

\bibitem{muglikar2023event}
Muglikar, M., Bauersfeld, L., Moeys, D.P., Scaramuzza, D.: Event-based shape from polarization. In: Proceedings of the IEEE/CVF Conference on Computer Vision and Pattern Recognition. pp. 1547--1556 (2023)

\bibitem{10161164}
Nagaraj, M., Liyanagedera, C.M., Roy, K.: Dotie - detecting objects through temporal isolation of events using a spiking architecture. In: 2023 IEEE International Conference on Robotics and Automation (ICRA). pp. 4858--4864 (2023). \doi{10.1109/ICRA48891.2023.10161164}

\bibitem{nagata2020qr}
Nagata, J., Sekikawa, Y., Hara, K., Suzuki, T., Aoki, Y.: Qr-code reconstruction from event data via optimization in code subspace. In: Proceedings of the IEEE/CVF Winter Conference on Applications of Computer Vision. pp. 2124--2132 (2020)

\bibitem{nam2022stereo}
Nam, Y., Mostafavi, M., Yoon, K.J., Choi, J.: Stereo depth from events cameras: Concentrate and focus on the future. In: Proceedings of the IEEE/CVF Conference on Computer Vision and Pattern Recognition. pp. 6114--6123 (2022)

\bibitem{10160967}
Ng, M., Cai, X., Foong, S.: Direct angular rate estimation without event motion-compensation at high angular rates. In: 2023 IEEE International Conference on Robotics and Automation (ICRA). pp. 1976--1981 (2023). \doi{10.1109/ICRA48891.2023.10160967}

\bibitem{ni2012asynchronous}
Ni, Z., Pacoret, C., Benosman, R., Ieng, S., R{\'E}GNIER*, S.: Asynchronous event-based high speed vision for microparticle tracking. Journal of microscopy  \textbf{245}(3),  236--244 (2012)

\bibitem{9812430}
Nunes, U.M., Demiris, Y.: Kinematic structure estimation of arbitrary articulated rigid objects for event cameras. In: 2022 International Conference on Robotics and Automation (ICRA). pp. 508--514 (2022). \doi{10.1109/ICRA46639.2022.9812430}

\bibitem{nunes2023time}
Nunes, U.M., Perrinet, L.U., Ieng, S.H.: Time-to-contact map by joint estimation of up-to-scale inverse depth and global motion using a single event camera. In: Proceedings of the IEEE/CVF International Conference on Computer Vision. pp. 23653--23663 (2023)

\bibitem{orchard2015converting}
Orchard, G., Jayawant, A., Cohen, G.K., Thakor, N.: Converting static image datasets to spiking neuromorphic datasets using saccades. Frontiers in neuroscience  \textbf{9}, ~437 (2015)

\bibitem{osswald2017spiking}
Osswald, M., Ieng, S.H., Benosman, R., Indiveri, G.: A spiking neural network model of 3d perception for event-based neuromorphic stereo vision systems. Scientific reports  \textbf{7}(1),  40703 (2017)

\bibitem{9523155}
Paikin, G., Ater, Y., Shaul, R., Soloveichik, E.: Efi-net: Video frame interpolation from fusion of events and frames. In: 2021 IEEE/CVF Conference on Computer Vision and Pattern Recognition Workshops (CVPRW). pp. 1291--1301 (2021). \doi{10.1109/CVPRW53098.2021.00142}

\bibitem{9252186}
Pan, L., Hartley, R., Scheerlinck, C., Liu, M., Yu, X., Dai, Y.: High frame rate video reconstruction based on an event camera. IEEE Transactions on Pattern Analysis and Machine Intelligence  \textbf{44}(5),  2519--2533 (2022). \doi{10.1109/TPAMI.2020.3036667}

\bibitem{pan2020single}
Pan, L., Liu, M., Hartley, R.: Single image optical flow estimation with an event camera. In: 2020 IEEE/CVF Conference on Computer Vision and Pattern Recognition (CVPR). pp. 1669--1678. IEEE (2020)

\bibitem{pan2019bringing}
Pan, L., Scheerlinck, C., Yu, X., Hartley, R., Liu, M., Dai, Y.: Bringing a blurry frame alive at high frame-rate with an event camera. In: Proceedings of the IEEE/CVF Conference on Computer Vision and Pattern Recognition. pp. 6820--6829 (2019)

\bibitem{parameshwara20210}
Parameshwara, C.M., Sanket, N.J., Singh, C.D., Ferm{\"u}ller, C., Aloimonos, Y.: 0-mms: Zero-shot multi-motion segmentation with a monocular event camera. In: 2021 IEEE International Conference on Robotics and Automation (ICRA). pp. 9594--9600. IEEE (2021)

\bibitem{paredes2021back}
Paredes-Vall{\'e}s, F., De~Croon, G.C.: Back to event basics: Self-supervised learning of image reconstruction for event cameras via photometric constancy. In: Proceedings of the IEEE/CVF Conference on Computer Vision and Pattern Recognition. pp. 3446--3455 (2021)

\bibitem{paredes2023taming}
Paredes-Vall{\'e}s, F., Scheper, K.Y., De~Wagter, C., De~Croon, G.C.: Taming contrast maximization for learning sequential, low-latency, event-based optical flow. In: Proceedings of the IEEE/CVF International Conference on Computer Vision. pp. 9695--9705 (2023)

\bibitem{peng2020globally}
Peng, X., Wang, Y., Gao, L., Kneip, L.: Globally-optimal event camera motion estimation. In: Computer Vision--ECCV 2020: 16th European Conference, Glasgow, UK, August 23--28, 2020, Proceedings, Part XXVI 16. pp. 51--67. Springer (2020)

\bibitem{peng2024scene}
Peng, Y., Li, H., Zhang, Y., Sun, X., Wu, F.: Scene adaptive sparse transformer for event-based object detection. In: Proceedings of the IEEE/CVF Conference on Computer Vision and Pattern Recognition. pp. 16794--16804 (2024)

\bibitem{GET_ICCV_2023}
Peng, Y., Zhang, Y., Xiong, Z., Sun, X., Wu, F.: Get: Group event transformer for event-based vision. In: 2023 IEEE/CVF International Conference on Computer Vision (ICCV). pp. 6015--6025 (2023). \doi{10.1109/ICCV51070.2023.00555}

\bibitem{perot2020learning}
Perot, E., De~Tournemire, P., Nitti, D., Masci, J., Sironi, A.: Learning to detect objects with a 1 megapixel event camera. Advances in Neural Information Processing Systems  \textbf{33},  16639--16652 (2020)

\bibitem{plizzari2022e2}
Plizzari, C., Planamente, M., Goletto, G., Cannici, M., Gusso, E., Matteucci, M., Caputo, B.: E2 (go) motion: Motion augmented event stream for egocentric action recognition. In: Proceedings of the IEEE/CVF conference on computer vision and pattern recognition. pp. 19935--19947 (2022)

\bibitem{ponghiran2023event}
Ponghiran, W., Liyanagedera, C.M., Roy, K.: Event-based temporally dense optical flow estimation with sequential learning. In: Proceedings of the IEEE/CVF International Conference on Computer Vision. pp. 9827--9836 (2023)

\bibitem{posch2014retinomorphic}
Posch, C., Serrano-Gotarredona, T., Linares-Barranco, B., Delbruck, T.: Retinomorphic event-based vision sensors: bioinspired cameras with spiking output. Proceedings of the IEEE  \textbf{102}(10),  1470--1484 (2014)

\bibitem{prophesee_event_simulator2023}
Prophesee: Video to event simulator (2023), \url{https://docs.prophesee.ai/stable/samples/modules/core_ml/viz_video_to_event_simulator.html}, accessed: 2024-07-17

\bibitem{metavision_evk3_genx320}
Prophesee: Metavision evk3 – genx320 (2024), \url{https://www.prophesee.ai/evk-3-genx320-info/}, accessed: 2024-07-18

\bibitem{metavision_evk3_hd}
Prophesee: Metavision evk3 – hd (2024), \url{https://www.prophesee.ai/event-based-evk-3/}, accessed: 2024-07-18

\bibitem{metavision_evK4_HD}
Prophesee: Metavision evk4 – hd (2024), \url{https://www.prophesee.ai/event-camera-evk4/}, accessed: 2024-07-18

\bibitem{metavision_sdk}
Prophesee: Metavision sdk (2024), \url{https://docs.prophesee.ai/stable/index.html}, accessed: 2024-07-18

\bibitem{metavision_kria_kv260}
Prophesee: Metavision starter kit – amd kria kv260 (2024), \url{https://www.prophesee.ai/event-based-metavision-amd-kria-starter-kit/}, accessed: 2024-07-18

\bibitem{metavision_stm32f7}
Prophesee: Metavision starter kit – stm32f7 (genx320) (2024), \url{https://www.prophesee.ai/stm32-genx320-info/}, accessed: 2024-07-18

\bibitem{prophesee_website}
Prophesee: Prophesee | metavision for machines (2024), \url{https://www.prophesee.ai/}, accessed: 2024-07-18

\bibitem{imx636_hd}
Prophesee, Sony: Imx 636 hd (2024), \url{https://www.prophesee.ai/event-based-sensor-imx636-sony-prophesee/}, accessed: 2024-07-18

\bibitem{qi2023e2nerf}
Qi, Y., Zhu, L., Zhang, Y., Li, J.: E2nerf: Event enhanced neural radiance fields from blurry images. In: Proceedings of the IEEE/CVF International Conference on Computer Vision. pp. 13254--13264 (2023)

\bibitem{qu2024implicit}
Qu, D., Yan, C., Wang, D., Yin, J., Chen, Q., Xu, D., Zhang, Y., Zhao, B., Li, X.: Implicit event-rgbd neural slam. In: Proceedings of the IEEE/CVF Conference on Computer Vision and Pattern Recognition. pp. 19584--19594 (2024)

\bibitem{ramesh2018long}
Ramesh, B., Zhang, S., Lee, Z.W., Gao, Z., Orchard, G., Xiang, C.: Long-term object tracking with a moving event camera. In: Bmvc. p.~241 (2018)

\bibitem{rebecq2018esim}
Rebecq, H., Gehrig, D., Scaramuzza, D.: Esim: an open event camera simulator. In: Conference on robot learning. pp. 969--982. PMLR (2018)

\bibitem{rebecq2019events}
Rebecq, H., Ranftl, R., Koltun, V., Scaramuzza, D.: Events-to-video: Bringing modern computer vision to event cameras. In: Proceedings of the IEEE/CVF Conference on Computer Vision and Pattern Recognition. pp. 3857--3866 (2019)

\bibitem{rebecq2019high}
Rebecq, H., Ranftl, R., Koltun, V., Scaramuzza, D.: High speed and high dynamic range video with an event camera. IEEE transactions on pattern analysis and machine intelligence  \textbf{43}(6),  1964--1980 (2019)

\bibitem{ren2024simple}
Ren, H., Zhu, J., Zhou, Y., Fu, H., Huang, Y., Cheng, B.: A simple and effective point-based network for event camera 6-dofs pose relocalization. In: Proceedings of the IEEE/CVF Conference on Computer Vision and Pattern Recognition. pp. 18112--18121 (2024)

\bibitem{rogister2011asynchronous}
Rogister, P., Benosman, R., Ieng, S.H., Lichtsteiner, P., Delbruck, T.: Asynchronous event-based binocular stereo matching. IEEE Transactions on Neural Networks and Learning Systems  \textbf{23}(2),  347--353 (2011)

\bibitem{rudnev2023eventnerf}
Rudnev, V., Elgharib, M., Theobalt, C., Golyanik, V.: Eventnerf: Neural radiance fields from a single colour event camera. In: Proceedings of the IEEE/CVF Conference on Computer Vision and Pattern Recognition. pp. 4992--5002 (2023)

\bibitem{rudneveventhands}
Rudnev, V., Golyanik, V., Wang, J., Seidel, H.P., Mueller, F., Elgharib, M., Theobalt, C.: Eventhands: Real-time neural 3d hand pose estimation from an event stream--supplementary document--

\bibitem{safa2022neuromorphic}
Safa, A., Van~Assche, J., Alea, M.D., Catthoor, F., Gielen, G.G.: Neuromorphic near-sensor computing: from event-based sensing to edge learning. Ieee Micro  \textbf{42}(6),  88--95 (2022)

\bibitem{10160681}
Safa, A., Verbelen, T., Ocket, I., Bourdoux, A., Sahli, H., Catthoor, F., Gielen, G.: Fusing event-based camera and radar for slam using spiking neural networks with continual stdp learning. In: 2023 IEEE International Conference on Robotics and Automation (ICRA). pp. 2782--2788 (2023). \doi{10.1109/ICRA48891.2023.10160681}

\bibitem{salvatore2022learned}
Salvatore, N., Fletcher, J.: Learned event-based visual perception for improved space object detection. In: Proceedings of the IEEE/CVF Winter Conference on Applications of Computer Vision. pp. 2888--2897 (2022)

\bibitem{schaefer2022aegnn}
Schaefer, S., Gehrig, D., Scaramuzza, D.: Aegnn: Asynchronous event-based graph neural networks. In: Proceedings of the IEEE/CVF conference on computer vision and pattern recognition. pp. 12371--12381 (2022)

\bibitem{Schaefer22cvpr_aegnn}
Schaefer, S., Gehrig, D., Scaramuzza, D.: Aegnn: Asynchronous event-based graph neural networks. In: IEEE Conference on Computer Vision and Pattern Recognition (2022)

\bibitem{scheerlinck2020fast}
Scheerlinck, C., Rebecq, H., Gehrig, D., Barnes, N., Mahony, R., Scaramuzza, D.: Fast image reconstruction with an event camera. In: Proceedings of the IEEE/CVF Winter Conference on Applications of Computer Vision. pp. 156--163 (2020)

\bibitem{schraml2010dynamic}
Schraml, S., Belbachir, A.N., Milosevic, N., Sch{\"o}n, P.: Dynamic stereo vision system for real-time tracking. In: Proceedings of 2010 IEEE International Symposium on Circuits and Systems. pp. 1409--1412. IEEE (2010)

\bibitem{10208126}
Sekikawa, Y., Nagata, J.: Live demonstration: Tangentially elongated gaussian belief propagation for event-based incremental optical flow estimation. In: 2023 IEEE/CVF Conference on Computer Vision and Pattern Recognition Workshops (CVPRW). pp. 3931--3932 (2023). \doi{10.1109/CVPRW59228.2023.00408}

\bibitem{v2e2023}
SensorsINI: v2e: A simulator for event-based vision (2023), \url{https://github.com/SensorsINI/v2e}, accessed: 2024-07-17

\bibitem{seok2020robust}
Seok, H., Lim, J.: Robust feature tracking in dvs event stream using b{\'e}zier mapping. In: Proceedings of the IEEE/CVF Winter Conference on Applications of Computer Vision. pp. 1658--1667 (2020)

\bibitem{serrano2015poker}
Serrano-Gotarredona, T., Linares-Barranco, B.: Poker-dvs and mnist-dvs. their history, how they were made, and other details. Frontiers in neuroscience  \textbf{9}, ~481 (2015)

\bibitem{shah2024codedevents}
Shah, S., Chan, M.A., Cai, H., Chen, J., Kulshrestha, S., Singh, C.D., Aloimonos, Y., Metzler, C.A.: Codedevents: Optimal point-spread-function engineering for 3d-tracking with event cameras. In: Proceedings of the IEEE/CVF Conference on Computer Vision and Pattern Recognition. pp. 25265--25275 (2024)

\bibitem{10160605}
Shi, D., Jing, L., Li, R., Liu, Z., Wang, L., Xu, H., Zhang, Y.: Improved event-based dense depth estimation via optical flow compensation. In: 2023 IEEE International Conference on Robotics and Automation (ICRA). pp. 4902--4908 (2023). \doi{10.1109/ICRA48891.2023.10160605}

\bibitem{shiba2022secrets}
Shiba, S., Aoki, Y., Gallego, G.: Secrets of event-based optical flow. In: European Conference on Computer Vision. pp. 628--645. Springer (2022)

\bibitem{Shiba24pami}
Shiba, S., Klose, Y., Aoki, Y., Gallego, G.: Secrets of event-based optical flow, depth, and ego-motion by contrast maximization. IEEE Trans. Pattern Anal. Mach. Intell. (T-PAMI) pp. 1--18 (2024). \doi{10.1109/TPAMI.2024.3396116}

\bibitem{carla_dvs2023}
Simulator, C.: Carla simulator dvs camera (2023), \url{https://carla.readthedocs.io/en/latest/ref_sensors/#dvs-camera}, accessed: 2024-07-17

\bibitem{sironi2018hats}
Sironi, A., Brambilla, M., Bourdis, N., Lagorce, X., Benosman, R.: Hats: Histograms of averaged time surfaces for robust event-based object classification. In: Proceedings of the IEEE conference on computer vision and pattern recognition. pp. 1731--1740 (2018)

\bibitem{imx636}
Sony, Prophesee: Imx 636 event-based sensor (2024), \url{https://www.prophesee.ai/event-based-sensor-imx636-sony-prophesee/}, accessed: 2024-07-18

\bibitem{van2002biologically}
Van~der Spiegel, J., Etienne-Cummings, R., Nishimura, M.: Biologically inspired vision sensors. In: 2002 23rd International Conference on Microelectronics. Proceedings (Cat. No. 02TH8595). vol.~1, pp. 125--131. IEEE (2002)

\bibitem{sridharan2024ev}
Sridharan, S., Selvam, S., Roy, K., Raghunathan, A.: Ev-edge: Efficient execution of event-based vision algorithms on commodity edge platforms. arXiv preprint arXiv:2403.15717  (2024)

\bibitem{stoffregen2019event}
Stoffregen, T., Gallego, G., Drummond, T., Kleeman, L., Scaramuzza, D.: Event-based motion segmentation by motion compensation. In: Proceedings of the IEEE/CVF International Conference on Computer Vision. pp. 7244--7253 (2019)

\bibitem{stromatias2017event}
Stromatias, E., Soto, M., Serrano-Gotarredona, T., Linares-Barranco, B.: An event-driven classifier for spiking neural networks fed with synthetic or dynamic vision sensor data. Frontiers in neuroscience  \textbf{11}, ~350 (2017)

\bibitem{sun2022event}
Sun, L., Sakaridis, C., Liang, J., Jiang, Q., Yang, K., Sun, P., Ye, Y., Wang, K., Gool, L.V.: Event-based fusion for motion deblurring with cross-modal attention. In: European conference on computer vision. pp. 412--428. Springer (2022)

\bibitem{sun2023event}
Sun, L., Sakaridis, C., Liang, J., Sun, P., Cao, J., Zhang, K., Jiang, Q., Wang, K., Van~Gool, L.: Event-based frame interpolation with ad-hoc deblurring. In: Proceedings of the IEEE/CVF Conference on Computer Vision and Pattern Recognition. pp. 18043--18052 (2023)

\bibitem{sun2022ess}
Sun, Z., Messikommer, N., Gehrig, D., Scaramuzza, D.: Ess: Learning event-based semantic segmentation from still images. In: European Conference on Computer Vision. pp. 341--357. Springer (2022)

\bibitem{iebcs2023}
Systems, N.: Iebcs: Icns event based camera simulator (2023), \url{https://github.com/neuromorphicsystems/IEBCS}, accessed: 2024-07-17

\bibitem{10161220}
Ta, K., Bruggemann, D., Brödermann, T., Sakaridis, C., Van~Gool, L.: L2e: Lasers to events for 6-dof extrinsic calibration of lidars and event cameras. In: 2023 IEEE International Conference on Robotics and Automation (ICRA). pp. 11425--11431 (2023). \doi{10.1109/ICRA48891.2023.10161220}

\bibitem{9879993}
Tan, G., Wang, Y., Han, H., Cao, Y., Wu, F., Zha, Z.J.: Multi-grained spatio-temporal features perceived network for event-based lip-reading. In: 2022 IEEE/CVF Conference on Computer Vision and Pattern Recognition (CVPR). pp. 20062--20071 (2022). \doi{10.1109/CVPR52688.2022.01946}

\bibitem{tan2022multi}
Tan, G., Wang, Y., Han, H., Cao, Y., Wu, F., Zha, Z.J.: Multi-grained spatio-temporal features perceived network for event-based lip-reading. In: Proceedings of the IEEE/CVF Conference on Computer Vision and Pattern Recognition. pp. 20094--20103 (2022)

\bibitem{9812059}
Tomy, A., Paigwar, A., Mann, K.S., Renzaglia, A., Laugier, C.: Fusing event-based and rgb camera for robust object detection in adverse conditions. In: 2022 International Conference on Robotics and Automation (ICRA). pp. 933--939 (2022). \doi{10.1109/ICRA46639.2022.9812059}

\bibitem{tulyakov2022time}
Tulyakov, S., Bochicchio, A., Gehrig, D., Georgoulis, S., Li, Y., Scaramuzza, D.: Time lens++: Event-based frame interpolation with parametric non-linear flow and multi-scale fusion. In: Proceedings of the IEEE/CVF Conference on Computer Vision and Pattern Recognition. pp. 17755--17764 (2022)

\bibitem{tulyakov2019learning}
Tulyakov, S., Fleuret, F., Kiefel, M., Gehler, P., Hirsch, M.: Learning an event sequence embedding for dense event-based deep stereo. In: Proceedings of the IEEE/CVF International Conference on Computer Vision. pp. 1527--1537 (2019)

\bibitem{tulyakov2021time}
Tulyakov, S., Gehrig, D., Georgoulis, S., Erbach, J., Gehrig, M., Li, Y., Scaramuzza, D.: Time lens: Event-based video frame interpolation. In: Proceedings of the IEEE/CVF conference on computer vision and pattern recognition. pp. 16155--16164 (2021)

\bibitem{verma2024etram}
Verma, A.A., Chakravarthi, B., Vaghela, A., Wei, H., Yang, Y.: etram: Event-based traffic monitoring dataset. In: Proceedings of the IEEE/CVF Conference on Computer Vision and Pattern Recognition. pp. 22637--22646 (2024)

\bibitem{vogelstein2007multichip}
Vogelstein, R.J., Mallik, U., Culurciello, E., Cauwenberghs, G., Etienne-Cummings, R.: A multichip neuromorphic system for spike-based visual information processing. Neural computation  \textbf{19}(9),  2281--2300 (2007)

\bibitem{wan2022s2n}
Wan, Z., Wang, Y., Tan, G., Cao, Y., Zha, Z.J.: S2n: Suppression-strengthen network for event-based recognition under variant illuminations. In: European Conference on Computer Vision. pp. 716--733. Springer (2022)

\bibitem{10376957}
Wan, Z., Mao, Y., Zhang, J., Dai, Y.: Rpeflow: Multimodal fusion of rgb-pointcloud-event for joint optical flow and scene flow estimation. In: 2023 IEEE/CVF International Conference on Computer Vision (ICCV). pp. 9996--10006 (2023). \doi{10.1109/ICCV51070.2023.00920}

\bibitem{wang2023unsupervised}
Wang, J., Weng, W., Zhang, Y., Xiong, Z.: Unsupervised video deraining with an event camera. In: Proceedings of the IEEE/CVF International Conference on Computer Vision. pp. 10831--10840 (2023)

\bibitem{wang2021evdistill}
Wang, L., Chae, Y., Yoon, S.H., Kim, T.K., Yoon, K.J.: Evdistill: Asynchronous events to end-task learning via bidirectional reconstruction-guided cross-modal knowledge distillation. In: Proceedings of the IEEE/CVF Conference on Computer Vision and Pattern Recognition. pp. 608--619 (2021)

\bibitem{wang2019event}
Wang, L., Ho, Y.S., Yoon, K.J., et~al.: Event-based high dynamic range image and very high frame rate video generation using conditional generative adversarial networks. In: Proceedings of the IEEE/CVF Conference on Computer Vision and Pattern Recognition. pp. 10081--10090 (2019)

\bibitem{wang2020eventsr}
Wang, L., Kim, T.K., Yoon, K.J.: Eventsr: From asynchronous events to image reconstruction, restoration, and super-resolution via end-to-end adversarial learning. In: Proceedings of the IEEE/CVF conference on computer vision and pattern recognition. pp. 8315--8325 (2020)

\bibitem{wang2019space}
Wang, Q., Zhang, Y., Yuan, J., Lu, Y.: Space-time event clouds for gesture recognition: From rgb cameras to event cameras. In: 2019 IEEE Winter Conference on Applications of Computer Vision (WACV). pp. 1826--1835. IEEE (2019)

\bibitem{wang2024event}
Wang, X., Wang, S., Tang, C., Zhu, L., Jiang, B., Tian, Y., Tang, J.: Event stream-based visual object tracking: A high-resolution benchmark dataset and a novel baseline. In: Proceedings of the IEEE/CVF Conference on Computer Vision and Pattern Recognition. pp. 19248--19257 (2024)

\bibitem{wang2019ev}
Wang, Y., Du, B., Shen, Y., Wu, K., Zhao, G., Sun, J., Wen, H.: Ev-gait: Event-based robust gait recognition using dynamic vision sensors. In: Proceedings of the IEEE/CVF conference on computer vision and pattern recognition. pp. 6358--6367 (2019)

\bibitem{wang2020stereo}
Wang, Y., Idoughi, R., Heidrich, W.: Stereo event-based particle tracking velocimetry for 3d fluid flow reconstruction. In: Computer Vision--ECCV 2020: 16th European Conference, Glasgow, UK, August 23--28, 2020, Proceedings, Part XXIX 16. pp. 36--53. Springer (2020)

\bibitem{wang2021asynchronous}
Wang, Z., Ng, Y., Scheerlinck, C., Mahony, R.: An asynchronous kalman filter for hybrid event cameras. In: Proceedings of the IEEE/CVF International Conference on Computer Vision. pp. 448--457 (2021)

\bibitem{9812003}
Wang, Z., Yuan, D., Ng, Y., Mahony, R.: A linear comb filter for event flicker removal. In: 2022 International Conference on Robotics and Automation (ICRA). pp. 398--404 (2022). \doi{10.1109/ICRA46639.2022.9812003}

\bibitem{weikersdorfer2014event}
Weikersdorfer, D., Adrian, D.B., Cremers, D., Conradt, J.: Event-based 3d slam with a depth-augmented dynamic vision sensor. In: 2014 IEEE international conference on robotics and automation (ICRA). pp. 359--364. IEEE (2014)

\bibitem{weikersdorfer2013simultaneous}
Weikersdorfer, D., Hoffmann, R., Conradt, J.: Simultaneous localization and mapping for event-based vision systems. In: Computer Vision Systems: 9th International Conference, ICVS 2013, St. Petersburg, Russia, July 16-18, 2013. Proceedings 9. pp. 133--142. Springer (2013)

\bibitem{weng2021event}
Weng, W., Zhang, Y., Xiong, Z.: Event-based video reconstruction using transformer. In: Proceedings of the IEEE/CVF International Conference on Computer Vision. pp. 2563--2572 (2021)

\bibitem{weng2023event}
Weng, W., Zhang, Y., Xiong, Z.: Event-based blurry frame interpolation under blind exposure. In: Proceedings of the IEEE/CVF Conference on Computer Vision and Pattern Recognition. pp. 1588--1598 (2023)

\bibitem{wu2022video}
Wu, S., You, K., He, W., Yang, C., Tian, Y., Wang, Y., Zhang, Z., Liao, J.: Video interpolation by event-driven anisotropic adjustment of optical flow. In: European Conference on Computer Vision. pp. 267--283. Springer (2022)

\bibitem{wu2024leod}
Wu, Z., Gehrig, M., Lyu, Q., Liu, X., Gilitschenski, I.: Leod: Label-efficient object detection for event cameras. In: Proceedings of the IEEE/CVF Conference on Computer Vision and Pattern Recognition. pp. 16933--16943 (2024)

\bibitem{xu2021motion}
Xu, F., Yu, L., Wang, B., Yang, W., Xia, G.S., Jia, X., Qiao, Z., Liu, J.: Motion deblurring with real events. In: Proceedings of the IEEE/CVF International Conference on Computer Vision. pp. 2583--2592 (2021)

\bibitem{xu2024dmr}
Xu, H., Peng, P., Tan, G., Li, Y., Xu, X., Tian, Y.: Dmr: Decomposed multi-modality representations for frames and events fusion in visual reinforcement learning. In: Proceedings of the IEEE/CVF Conference on Computer Vision and Pattern Recognition. pp. 26508--26518 (2024)

\bibitem{xu2020eventcap}
Xu, L., Xu, W., Golyanik, V., Habermann, M., Fang, L., Theobalt, C.: Eventcap: Monocular 3d capture of high-speed human motions using an event camera. In: Proceedings of the IEEE/CVF Conference on Computer Vision and Pattern Recognition. pp. 4968--4978 (2020)

\bibitem{yang2023learning}
Yang, Y., Han, J., Liang, J., Sato, I., Shi, B.: Learning event guided high dynamic range video reconstruction. In: Proceedings of the IEEE/CVF Conference on Computer Vision and Pattern Recognition. pp. 13924--13934 (2023)

\bibitem{yang2024latency}
Yang, Y., Liang, J., Yu, B., Chen, Y., Ren, J.S., Shi, B.: Latency correction for event-guided deblurring and frame interpolation. In: Proceedings of the IEEE/CVF Conference on Computer Vision and Pattern Recognition. pp. 24977--24986 (2024)

\bibitem{yao2021temporal}
Yao, M., Gao, H., Zhao, G., Wang, D., Lin, Y., Yang, Z., Li, G.: Temporal-wise attention spiking neural networks for event streams classification. In: Proceedings of the IEEE/CVF International Conference on Computer Vision. pp. 10221--10230 (2021)

\bibitem{9278914}
Yasin, J.N., Mohamed, S.A.S., Haghbayan, M.h., Heikkonen, J., Tenhunen, H., Yasin, M.M., Plosila, J.: Night vision obstacle detection and avoidance based on bio-inspired vision sensors. In: 2020 IEEE SENSORS. pp.~1--4 (2020). \doi{10.1109/SENSORS47125.2020.9278914}

\bibitem{ye2020unsupervised}
Ye, C., Mitrokhin, A., Ferm{\"u}ller, C., Yorke, J.A., Aloimonos, Y.: Unsupervised learning of dense optical flow, depth and egomotion with event-based sensors. In: 2020 IEEE/RSJ International Conference on Intelligent Robots and Systems (IROS). pp. 5831--5838. IEEE (2020)

\bibitem{yu2024eventps}
Yu, B., Ren, J., Han, J., Wang, F., Liang, J., Shi, B.: Eventps: Real-time photometric stereo using an event camera. In: Proceedings of the IEEE/CVF Conference on Computer Vision and Pattern Recognition. pp. 9602--9611 (2024)

\bibitem{yu2021training}
Yu, Z., Zhang, Y., Liu, D., Zou, D., Chen, X., Liu, Y., Ren, J.S.: Training weakly supervised video frame interpolation with events. In: Proceedings of the IEEE/CVF International Conference on Computer Vision. pp. 14589--14598 (2021)

\bibitem{zhang2022data}
Zhang, D., Ding, Q., Duan, P., Zhou, C., Shi, B.: Data association between event streams and intensity frames under diverse baselines. In: European Conference on Computer Vision. pp. 72--90. Springer (2022)

\bibitem{zhang2022spiking}
Zhang, J., Dong, B., Zhang, H., Ding, J., Heide, F., Yin, B., Yang, X.: Spiking transformers for event-based single object tracking. In: Proceedings of the IEEE/CVF conference on Computer Vision and Pattern Recognition. pp. 8801--8810 (2022)

\bibitem{zhang2023frame}
Zhang, J., Wang, Y., Liu, W., Li, M., Bai, J., Yin, B., Yang, X.: Frame-event alignment and fusion network for high frame rate tracking. In: Proceedings of the IEEE/CVF Conference on Computer Vision and Pattern Recognition. pp. 9781--9790 (2023)

\bibitem{zhang2021object}
Zhang, J., Yang, X., Fu, Y., Wei, X., Yin, B., Dong, B.: Object tracking by jointly exploiting frame and event domain. In: Proceedings of the IEEE/CVF International Conference on Computer Vision. pp. 13043--13052 (2021)

\bibitem{zhang2022discrete}
Zhang, K., Che, K., Zhang, J., Cheng, J., Zhang, Z., Guo, Q., Leng, L.: Discrete time convolution for fast event-based stereo. In: Proceedings of the IEEE/CVF Conference on Computer Vision and Pattern Recognition. pp. 8676--8686 (2022)

\bibitem{10138453}
Zhang, P., Ge, Z., Song, L., Lam, E.Y.: Neuromorphic imaging with density-based spatiotemporal denoising. IEEE Transactions on Computational Imaging  \textbf{9},  530--541 (2023). \doi{10.1109/TCI.2023.3281202}

\bibitem{zhang2020learning}
Zhang, S., Zhang, Y., Jiang, Z., Zou, D., Ren, J., Zhou, B.: Learning to see in the dark with events. In: Computer Vision--ECCV 2020: 16th European Conference, Glasgow, UK, August 23--28, 2020, Proceedings, Part XVIII 16. pp. 666--682. Springer (2020)

\bibitem{zhang2022unifying}
Zhang, X., Yu, L.: Unifying motion deblurring and frame interpolation with events. In: Proceedings of the IEEE/CVF Conference on Computer Vision and Pattern Recognition. pp. 17765--17774 (2022)

\bibitem{zhang2023generalizing}
Zhang, X., Yu, L., Yang, W., Liu, J., Xia, G.S.: Generalizing event-based motion deblurring in real-world scenarios. In: Proceedings of the IEEE/CVF International Conference on Computer Vision. pp. 10734--10744 (2023)

\bibitem{zhang2023v2ce}
Zhang, Z., Cui, S., Chai, K., Yu, H., Dasgupta, S., Mahbub, U., Rahman, T.: V2ce: Video to continuous events simulator. arXiv preprint arXiv:2309.08891  (2023)

\bibitem{10160472}
Zhao, C., Li, Y., Lyu, Y.: Event-based real-time moving object detection based on imu ego-motion compensation. In: 2023 IEEE International Conference on Robotics and Automation (ICRA). pp. 690--696 (2023). \doi{10.1109/ICRA48891.2023.10160472}

\bibitem{zheng2024eventdance}
Zheng, X., Wang, L.: Eventdance: Unsupervised source-free cross-modal adaptation for event-based object recognition. In: Proceedings of the IEEE/CVF Conference on Computer Vision and Pattern Recognition. pp. 17448--17458 (2024)

\bibitem{zhou2024bring}
Zhou, H., Chang, Y., Shi, Z.: Bring event into rgb and lidar: Hierarchical visual-motion fusion for scene flow. In: Proceedings of the IEEE/CVF Conference on Computer Vision and Pattern Recognition. pp. 26477--26486 (2024)

\bibitem{zhou2024exact}
Zhou, J., Zheng, X., Lyu, Y., Wang, L.: Exact: Language-guided conceptual reasoning and uncertainty estimation for event-based action recognition and more. In: Proceedings of the IEEE/CVF Conference on Computer Vision and Pattern Recognition. pp. 18633--18643 (2024)

\bibitem{EvUnroll}
Zhou, X., Duan, P., Ma, Y., Shi, B.: Evunroll: Neuromorphic events based rolling shutter image correction. In: 2022 IEEE/CVF Conference on Computer Vision and Pattern Recognition (CVPR). pp. 17754--17763 (2022). \doi{10.1109/CVPR52688.2022.01725}

\bibitem{zhou2018semi}
Zhou, Y., Gallego, G., Rebecq, H., Kneip, L., Li, H., Scaramuzza, D.: Semi-dense 3d reconstruction with a stereo event camera. In: Proceedings of the European conference on computer vision (ECCV). pp. 235--251 (2018)

\bibitem{zhou2021event}
Zhou, Y., Gallego, G., Shen, S.: Event-based stereo visual odometry. IEEE Transactions on Robotics  \textbf{37}(5),  1433--1450 (2021)

\bibitem{10161563}
Zhou, Z., Wu, Z., Boutteau, R., Yang, F., Demonceaux, C., Ginhac, D.: Rgb-event fusion for moving object detection in autonomous driving. In: 2023 IEEE International Conference on Robotics and Automation (ICRA). pp. 7808--7815 (2023). \doi{10.1109/ICRA48891.2023.10161563}

\bibitem{zhu2018multivehicle}
Zhu, A.Z., Thakur, D., {\"O}zaslan, T., Pfrommer, B., Kumar, V., Daniilidis, K.: The multivehicle stereo event camera dataset: An event camera dataset for 3d perception. IEEE Robotics and Automation Letters  \textbf{3}(3),  2032--2039 (2018)

\bibitem{zhu2018ev}
Zhu, A.Z., Yuan, L., Chaney, K., Daniilidis, K.: Ev-flownet: Self-supervised optical flow estimation for event-based cameras. arXiv preprint arXiv:1802.06898  (2018)

\bibitem{zhu2019unsupervised}
Zhu, A.Z., Yuan, L., Chaney, K., Daniilidis, K.: Unsupervised event-based learning of optical flow, depth, and egomotion. In: Proceedings of the IEEE/CVF Conference on Computer Vision and Pattern Recognition. pp. 989--997 (2019)

\bibitem{zhu2022event}
Zhu, L., Wang, X., Chang, Y., Li, J., Huang, T., Tian, Y.: Event-based video reconstruction via potential-assisted spiking neural network. In: Proceedings of the IEEE/CVF Conference on Computer Vision and Pattern Recognition. pp. 3594--3604 (2022)

\bibitem{zhu2023cross}
Zhu, Z., Hou, J., Wu, D.O.: Cross-modal orthogonal high-rank augmentation for rgb-event transformer-trackers. In: Proceedings of the IEEE/CVF International Conference on Computer Vision. pp. 22045--22055 (2023)

\bibitem{10160654}
Ziegler, A., Teigland, D., Tebbe, J., Gossard, T., Zell, A.: Real-time event simulation with frame-based cameras. In: 2023 IEEE International Conference on Robotics and Automation (ICRA). pp. 11669--11675 (2023). \doi{10.1109/ICRA48891.2023.10160654}

\bibitem{zihao2017event}
Zihao~Zhu, A., Atanasov, N., Daniilidis, K.: Event-based visual inertial odometry. In: Proceedings of the IEEE Conference on Computer Vision and Pattern Recognition. pp. 5391--5399 (2017)

\bibitem{zou2021eventhpe}
Zou, S., Guo, C., Zuo, X., Wang, S., Wang, P., Hu, X., Chen, S., Gong, M., Cheng, L.: Eventhpe: Event-based 3d human pose and shape estimation. In: Proceedings of the IEEE/CVF International Conference on Computer Vision. pp. 10996--11005 (2021)

\bibitem{zou2021learning}
Zou, Y., Zheng, Y., Takatani, T., Fu, Y.: Learning to reconstruct high speed and high dynamic range videos from events. In: Proceedings of the IEEE/CVF Conference on Computer Vision and Pattern Recognition. pp. 2024--2033 (2021)

\bibitem{zubic2023chaos}
Zubi{\'c}, N., Gehrig, D., Gehrig, M., Scaramuzza, D.: From chaos comes order: Ordering event representations for object recognition and detection. In: Proceedings of the IEEE/CVF International Conference on Computer Vision. pp. 12846--12856 (2023)

\bibitem{Zubic_2024_CVPR_SSM}
Zubi\'c, N., Gehrig, M., Scaramuzza, D.: State space models for event cameras. In: Proceedings of the IEEE/CVF Conference on Computer Vision and Pattern Recognition (CVPR) (2024)

\bibitem{9811805}
Zuo, Y., Yang, J., Chen, J., Wang, X., Wang, Y., Kneip, L.: Devo: Depth-event camera visual odometry in challenging conditions. In: 2022 International Conference on Robotics and Automation (ICRA). pp. 2179--2185 (2022). \doi{10.1109/ICRA46639.2022.9811805}

\end{thebibliography}
\end{document}